\documentclass[sigconf]{acmart}

\AtBeginDocument{%
  }

\copyrightyear{2024}
\acmYear{2024}
\setcopyright{licensedusgovmixed}\acmConference[GeoAI'24]{7th ACM SIGSPATIAL International Workshop on AI for Geographic Knowledge Discovery}{October 29-November 1 2024}{Atlanta, GA, USA}
\acmBooktitle{7th ACM SIGSPATIAL International Workshop on AI for Geographic Knowledge Discovery (GeoAI'24), October 29-November 1 2024, Atlanta, GA, USA}
\acmDOI{10.1145/3687123.3698284}
\acmISBN{979-8-4007-1176-3/24/10}

\ccsdesc[500]{Computing methodologies~Neural networks}
\ccsdesc[500]{Computing methodologies~Spatial and physical reasoning}

\usepackage{subfig,multirow}
\usepackage{enumitem}

\begin{document}

\title{Encoding Agent Trajectories as Representations with Sequence Transformers}

\author{Athanasios Tsiligkaridis}
\email{athanasios.tsiligkaridis@str.us}
\affiliation{%
  \institution{Systems \& Technology Research (STR)}
  \city{Woburn}
  \state{MA}
  \country{USA}
}

\author{Nicholas Kalinowski}
\email{nicholas.kalinowski@str.us}
\affiliation{%
  \institution{Systems \& Technology Research (STR)}
  \city{Arlington}
  \state{VA}
  \country{USA}
}

\author{Zhongheng Li}
\email{zhongheng.li@str.us}
\affiliation{%
  \institution{Systems \& Technology Research (STR)}
  \city{Woburn}
  \state{MA}
  \country{USA}
}
\author{Elizabeth Hou}
\email{elizabeth.hou@str.us}
\affiliation{%
  \institution{Systems \& Technology Research (STR)}
  \city{Arlington}
  \state{VA}
  \country{USA}
}

\begin{abstract}
Spatiotemporal data faces many analogous challenges to natural language text including the ordering of locations (words) in a sequence, long range dependencies between locations, and locations having multiple meanings. In this work, we propose a novel model for representing high dimensional spatiotemporal trajectories as sequences of discrete locations and encoding them with a Transformer-based neural network architecture. Similar to language models, our Sequence Transformer for Agent Representation Encodings (STARE) model can learn representations and structure in trajectory data through both supervisory tasks (e.g., classification), and self-supervisory tasks (e.g., masked modelling). We present experimental results on various synthetic and real trajectory datasets and show that our proposed model can learn meaningful encodings that are useful for many downstream tasks including discriminating between labels and indicating similarity between locations. Using these encodings, we also learn relationships between agents and locations present in spatiotemporal data.  
\end{abstract}

\keywords{Transformers, encoders, spatiotemporal data, human mobility, trajectory modeling}
\maketitle

\section{Introduction}

In the modern world, geospatial mobility data has become increasingly available with the proliferation of mobile devices and other positioning and sensor technologies \cite{geolife, japan_data}. These technologies have also allowed wildlife researchers to monitor and collect data on animal movements and their ecosystems \cite{seabird_data, movebank_data}. This increasing availability of location datasets has been leveraged by researchers allowing them to build models that further their understanding of mobility patterns \cite{traffic, migration}, trafficking, and infectious disease \cite{covid1, birdflu} (in both humans and animals). Inspired by the successes of neural networks in other fields (e.g., vision, language), researchers have leveraged their flexible architecture and applied them to trajectory data to model human and animal mobility patterns \cite{dl_survey, animal_ml}. 

The sequential nature of trajectory data shares many similar properties with the sequential nature of natural language. Tokenizing sentences or text segments in Natural Language Processing (NLP) is essentially a mapping from a complex ``quasi-infinite" space to a sequence of discrete elements in a finite vocabulary. In a similar fashion to rule-based or algorithmic tokenizers \cite{spacy, bpe}, the ``quasi-continuous" high frequency GPS coordinates in trajectory data can also be tokenized into a lower dimensional sequence with a finite vocabulary. Also, just as tokens in language settings have semantic meaning (e.g., $[fir]$ refers to a coniferous tree), tokens in mobility data can also have inherent meaning about the location they represent (e.g., $(45.832119, 6.865575)$ is in the French alps). The order of the tokens in both domains also contains key information as language tokens can modify each other and location tokens can have different implicit meanings depending on their order (e.g., visiting a supermarket before work makes it a breakfast location whereas after work makes it a grocery shop). We can further extend our analogy between NLP and trajectory data to the ``document" level where just as sentences in the same document will share similar properties such as word choice, grammar, and style, trajectories collected from the same agent or user will also share similar orderings of locations or routes. Thus, we leverage these numerous similarities by proposing the Sequence Transformer for Agent Representation Encodings (STARE) model, a neural network with an encoder-based transformer architecture similar to those used in BERT-like language models \cite{devlin2018bert, vaswani2017attention}.

\subsection{Related Work}

Other works have also observed the connection between the sequential nature of trajectory data and language or other sequential domains. Transformers and other sequence deep-learning based approaches have been explored as potential frameworks for various trajectory-related tasks. Earlier works, \cite{Brebisson:2015, Xu:2018, Shi:2019}, focused on more simple architectures, such as RNN and LSTM models, which were successfully employed to solve destination and trajectory prediction tasks. 

As improvements to these models, transformer-based architectures emerged due to their robust attention mechanisms that allow for sequence dependency modeling without consideration of element distances along with their ability to be efficiently trained. Thus, later works leveraged the power of transformer-style architectures for a wide range of geospatial-related tasks. \cite{Tsiligkaridis:2020, Giuliari:2021, Abideen:2021} focused on forecasting trajectories for next destination prediction, while \cite{Quinatar:2021} augmented information and incorporated it into a transformer network to predict vehicle trajectories in urban scenarios. A transformer decoder-based architecture was proposed to predict next intended locations of users given contextual information \cite{Ye:2022}, a spatio-temporal contextual transformer fused with external features was used to segment imperfect historical maps \cite{Wu:2023}, and a graph transformer was used for point of interest recommendation and mobility modeling \cite{Xu:2023}. The trajectory recovery task was studied using a road network graph-aware transformer-based model for capturing spatiotemporal information from low sample trajectories and a multi-task decoder model for point generation \cite{Chen:2023}. A unified framework based on BERT has been explored as a means of solving a plethora of trajectory-related problems, such as imputation, classification, prediction, etc. \cite{Mashaal:2022}. Some of these transformer-based architectures also added significant amounts of side information, in the form of a contextual block \cite{Tsiligkaridis:2022} or multiple feature extractors that incorporate a points of interest ontology and a social network \cite{MobTCast}. Instead of just having a similar architecture, \cite{spabert} and \cite{MobTCast2} extended pre-trained large language models to take in pseudo-sentences that represent locations in natural language, effectively performing an NLP task. 

In contrast to the architecture presented in \cite{Tsiligkaridis:2022}, our approach solely requires time-stamped spatiotemporal data, carries out a different form of pre-processing, and considers various training regimes (i.e., label classification, masked token modeling). In contrast to the architecture presented in \cite{Zhou:2022} which has separate spatial and temporal transformers and a cross attention module between them, we have spatio-temporal transformers to allow for more shared information between the space and time domains. With this, our STARE model has the flexibility to be used in a variety of settings where only location and time data is present, such as with various animal trajectory datasets and in environments without contextual and/or foundational information. Our model simply takes in \textit{only} spatiotemporal data and ultimately learns important encodings that can be leveraged for various downstream tasks (e.g., classification, destination prediction, clustering) and for learning recurring behaviors, relationships between, and Patterns of Life (PoL) of various agents of interest.  

\subsection{Contributions Of Our Work}

In this paper, we make the following contributions.  First, we present a data discretization technique for reducing the dimensionality of long and rich agent PoL data.  Second, we propose a novel transformer-based architecture for obtaining informative data embeddings which can be used to learn relationships between agents and locations.  Finally, we present extensive experiments on both simulated and real trajectory datasets and showcase our proposed STARE model's informative embeddings along with improved performance over baseline sequence encoder models (i.e., LSTM, BiLSTM) in regards to classification accuracy. We also present novel experiments leveraging the transformer-based architecture of STARE to also learn the intrinsic patterns within the sequences themselves; specifically, we learn relationships between agents and between locations that these agents frequent.  %(which can blow up in size given a high data sampling rate and large timeframe for data collection) 

\section{STARE Model}

In this section, we describe our STARE model, which compresses raw trajectory data into novel tokenized sequences for input into a Transformer Encoder Stack (TES). Our architecture is similar to that of \cite{Tsiligkaridis:2022}, but we specifically focus on the minimal data setting where we solely have sequence information as data (i.e., we do not use a contextual block in our input as we do not incorporate any non-sequential information in our model). The aim of STARE is to discretize raw trajectory data for a TES to learn encodings that have semantic meaning and, in turn, can be used to both predict labels of interest with a Multi-Layer Perceptron (MLP) and also learn interesting relationships between observed data.

\subsection{Data Discretization Methodology} \label{subsec:preprocessing}

We begin with a dataset $\mathcal{X}$ containing $N$ observations of a tuple containing the latitude, longitude, and timestamp of each agent for $A$ total agents. To form multiple samples per agent and to reduce the size of our model architecture, we make independence assumptions on the temporal component by assuming that timepoints within a time window are dependent with each other, but independent of those outside the window. Explicitly, we partition $X_a$, the trajectory of agent $a$, into a set of $M$ sub-trajectories, \{$X_1$, $X_2$, \ldots, $X_M$\} where each sub-trajectory is the length of some time window (e.g. 6 hours, a day, or a week). The choice of the time window length is dataset dependent, but it uses the implicit assumption that repetitive behavior is expected and independent of each other. For example, humans tend to have cycles of the same behavior over days (e.g., people live in one place, wake up, go to work, do some activities throughout the day, and then return home to end their day) with some seasonality and anomalies. These independence assumptions are very typical in the sequential domain and parallel the breaking of text into multiple samples between sentences in NLP.  

For a given agent $a \in [1,...,A]$ and a time window $m \in [1,...,M]$, we define a trajectory $T_{a,m}$ as: 
\begin{description}[leftmargin=5pt, labelindent=0pt]
  \item[Definition 1 (Trajectory $T_{a,m}$)] For a time window $m$, agent $a$ has a raw trajectory $T_{a,m}$ that is defined as a sequence of $L_{a,m}$ time-stamped locations: $T_{a,m} = [p_1, \ldots, p_{L_{a,m}}]$, where each point $p_i = [\text{lat}_i,\text{lon}_i,t_i]$, is a tuple of latitude, longitude, and time, respectively, that identifies the geographic location of that point $p_i$ at time $t_i$.  
\end{description}

The number of data points in $T_{a,m}$ can be large and varying between different agents and time windows due to uneven measurement sampling rates and durations. Since locations are defined by their quasi-continuous latitude and longitude values, there is an infinitesimally large number of unique locations. To combat these issues, we discretize the data inputs by only retaining information about an agent's Persistent Locations (PL) and the times they spend in these PLs. 

Specifically, we define a PL as a stationary point in a trajectory that is mapped to a discrete value in an alphabet, where we define this alphabet to be S2 cells (a hierarchy of indexed spatial cells that represent geographical areas over the world) at a certain zoom level \cite{s2geometry}. We can also discretize the time spent in PLs by defining another alphabet that consists of multiples of some time measure, (e.g., minutes, hours) and rounding the PL times to the nearest multiple. By including the amount of time spent in a PL, we can represent a sequence of points whose latitude and longitude values lie in the same S2 cells in a more information-compact form (e.g. $[A, A, A, A] = {A, 4}$ where $A$ represents a S2 cell hash and $4$ is the amount of time spent in $A$ for the corresponding time measure). 

\begin{figure}[!h] 
    \centering
    \includegraphics[width=0.95\linewidth]{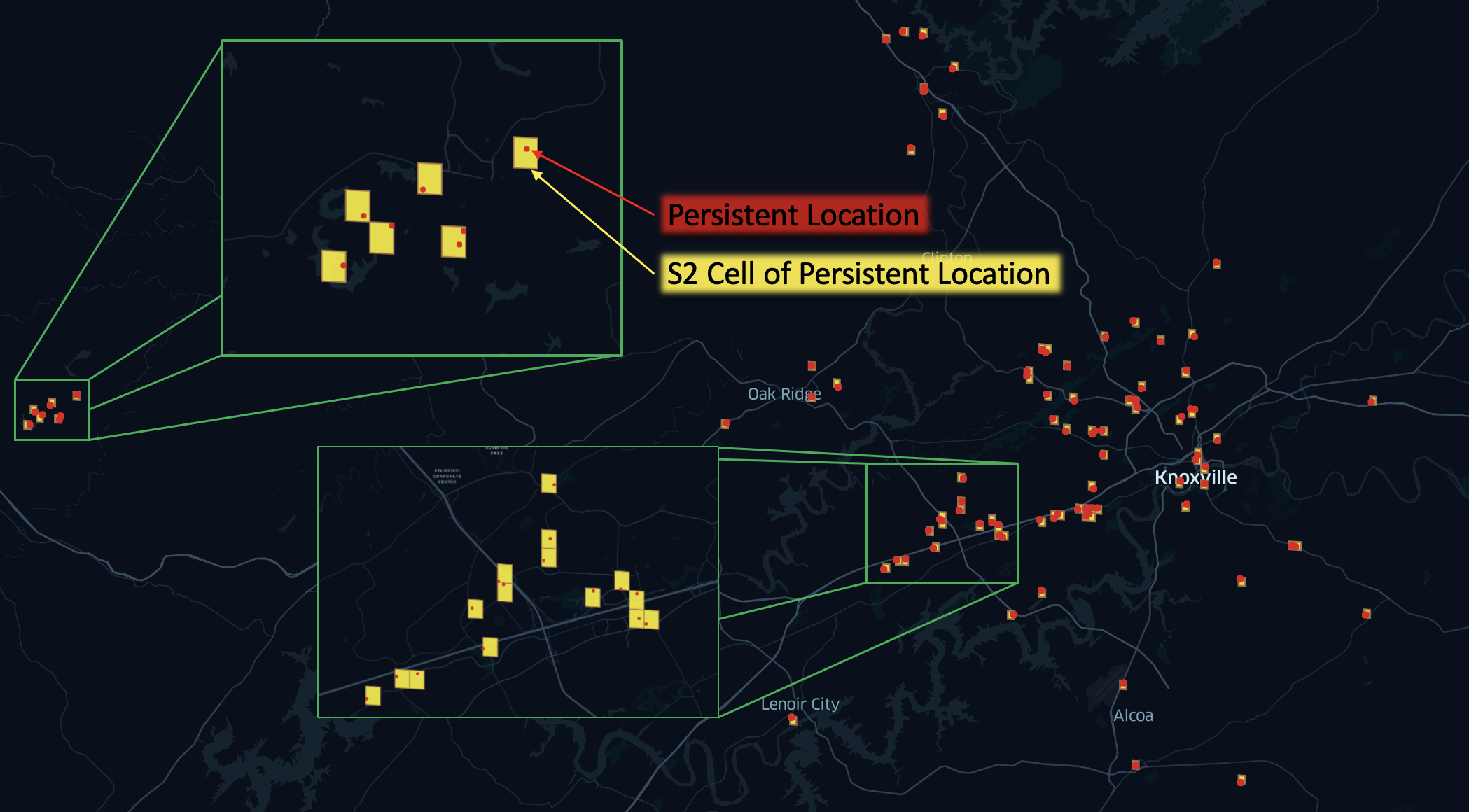}
    \caption{Example of extracted PLs (shown in red) from an agent's PoL along with the S2 cells (shown in yellow) that each PL resides in.}
    \label{fig:PL_visualize}
\end{figure}

Let $ s: [\Phi, \Lambda] \rightarrow \mathbb{S}^2$ be a mapping from latitude, longitude space to our S2 cell alphabet and $dt: \mathbb{R}^+ \rightarrow \mathbb{T}$ be a mapping from the positive real space of PL times to our discrete alphabet of rounded times. Thus, the concatenated sub-sequences of visited PLs and time spent in them form a sequence sample, $x_{a,m}$, that is a discretized and compressed version of $T_{a,m}$: 
\begin{flalign} 
x_{a,m} = \Big[[BOS], s(T_{a,m}), [SEP], dt(T_{a,m}), [EOS]\Big], \label{tok_seq}
\end{flalign}
where the $[BOS], [EOS], [SEP]$ tokens, borrowed from NLP, represent beginning, end, and separating tokens, respectively. Figure \ref{fig:PL_visualize} shows an example of how the PLs get mapped to S2 cells; unless the PLs are proximal, they are mapped to unique cells. We present an example of our data discretization method in Appendix \ref{subsec:example_discretization}.

Regarding the discretization of PLs, the choice of S2 zoom level affects spatial resolution.  A small zoom level yields large spatial cells where all the coordinates contained in a given cell are represented by that cell's index.  Equivalently, a larger zoom level yields smaller cells that can allow for more fine-grained representations of spatial areas.  A larger zoom level will yield a larger amount of total cells that are used to cover a given area, and vice versa.  The choice of S2 cell size is ultimately a task-dependent choice that can be selected through experimentation.  

For our problem, the choice of S2 discretization can also affect the performance of our models since we aim to learn and extract interesting characteristics and relationships of rich PoL data.  A small zoom level can yield a small amount of cells, and in turn a small vocabulary for our sequence model, but there could be a case where many obtained PLs in our data, that might represent different entities yet be fairly close in space (i.e., a hotel that is on the same street as a movie theater), are all grouped into a single S2 cell.  A large amount of this proximal PL \emph{blending} can result in a lack of identification and extraction of true relationships in data.
On the other end, using a large S2 cell zoom level can yield more S2 cells and the proximal PL blending situation can be minimized. With a larger set of relevant cells, a larger vocabulary will result, and more model training may be required for the model to fully understand the structure of the input data.  With more \emph{focused} S2 cells, we can expect more characteristics in data to be identified and extracted; however, we avoid using a very high zoom level as we do want some data compression to be carried out. 

In addition to S2 cell discretization, the general compression of the raw trajectories is a necessary modelling condition and has numerous benefits including:
\begin{enumerate}
\item Creating a manageable, finite size vocabulary instead of all latitude, longitude, and time values in the dataset
\item Reducing the length of the trajectories to be a function of only discrete stationary locations allowing for more manageable sized models
\item Allowing for self-supervised reconstruction learning (i.e. "masked language modelling") of the sequence due to the discrete, finite vocabulary
\end{enumerate}
Additionally, our data compression method is advantageous in situations where low sample data trajectories exist.  In some domain-specific datasets, such as some animal datasets where trajectories are not recorded with high confidence, rich and complete trajectories might not exist, but our approach of defining data sequences using visited PLs and duration times would remain valid as there would still be some points that would allow us to get correct estimates of PLs and approximate duration times at these PLs.  Other approaches that either require road network path information or do not use a data compression module would falter in this scenario.   

\subsection{STARE Model Architecture} \label{subsec:stare}

Evidently, our data discretization technique is a means of tokenizing our input to be passed into our Transformer-based model. Let $X$ be the inputs to the transformer model (i.e., our tokenized sequences split up according to some time window $m$), and let $Y$ be the encoded label associated with $X$ (e.g., the ID of the agent). The Transformer-based model then learns a non-linear embedding function $f(X)$ that produces an encoded representation of the sequence in continuous space $\mathbb{R}^{K}$ where $K$ is the dimensionality of this space. In order to provide a supervisory signal to train our model and learn this encoded representation, we have two training schemes or "task" heads to put on top of the transformer backbone function $f(X)$.  The use of two different training schemes allows us to learn various things about our data; specifically, the first task head allows us to learn relationships between agents while the second one allows us to learn relationships between locations that these agents visit.  

The first type of task head is a classifier $c(\cdot)$ (multi-layer perceptron) whose goal is to learn the conditional distribution $P(Y|X)$ which is the inverse of the traditional assumed data generating distribution $P(X|Y)$. We encode this information into one of the special tokens ($[BOS]$ is standard practice) and minimize the cross-entropy loss:
\begin{flalign} 
\min \mathcal{L}(Y, \hat{Y}) \quad \text{, where} \quad \hat{Y} = c(f(X)_{0}),
\end{flalign}
and where the first element of the encoded representation $f(X)_{0}$ is a $K$-dimensional vector and corresponds to the $[BOS]$ token. 

The second type of task head is a ``decoder" (linear layer without a bias term) that performs masked location modelling and learns the marginal distribution of sequences $P(X)$ by learning to predict the patterns with the sequences. Masked location modelling is a self-supervised training scheme where at every iteration, a random percentage of the inputs are masked as $X'$. The goal of the model is to learn encoded representations of the locations within each of the sequences such that they can be ``decoded". So for every masked token in the input sequence, the cross-entropy loss is calculated with respect to the un-masked token:
\begin{flalign} 
\min \mathcal{L}(X_i, \hat{X}_i) \quad \text{, where} \quad \hat{X} = d(f(X')_i) \quad \forall i \in Masked,
\end{flalign}
where $d(\cdot)$ is the decoder function and the $Masked$ set is a random subset of the location tokens. We do not mask the time tokens because while they provide supplementary information about the locations in the context of different agent PoLs, they do not have an intrinsic meaning on their own. We also do not mask the special tokens because they are meaningless; although we note that by encoding the classification representation in a special token, these two tasks are complementary and can be performed together. 

\begin{figure}[t]
    \centering
    \includegraphics[width=\linewidth]{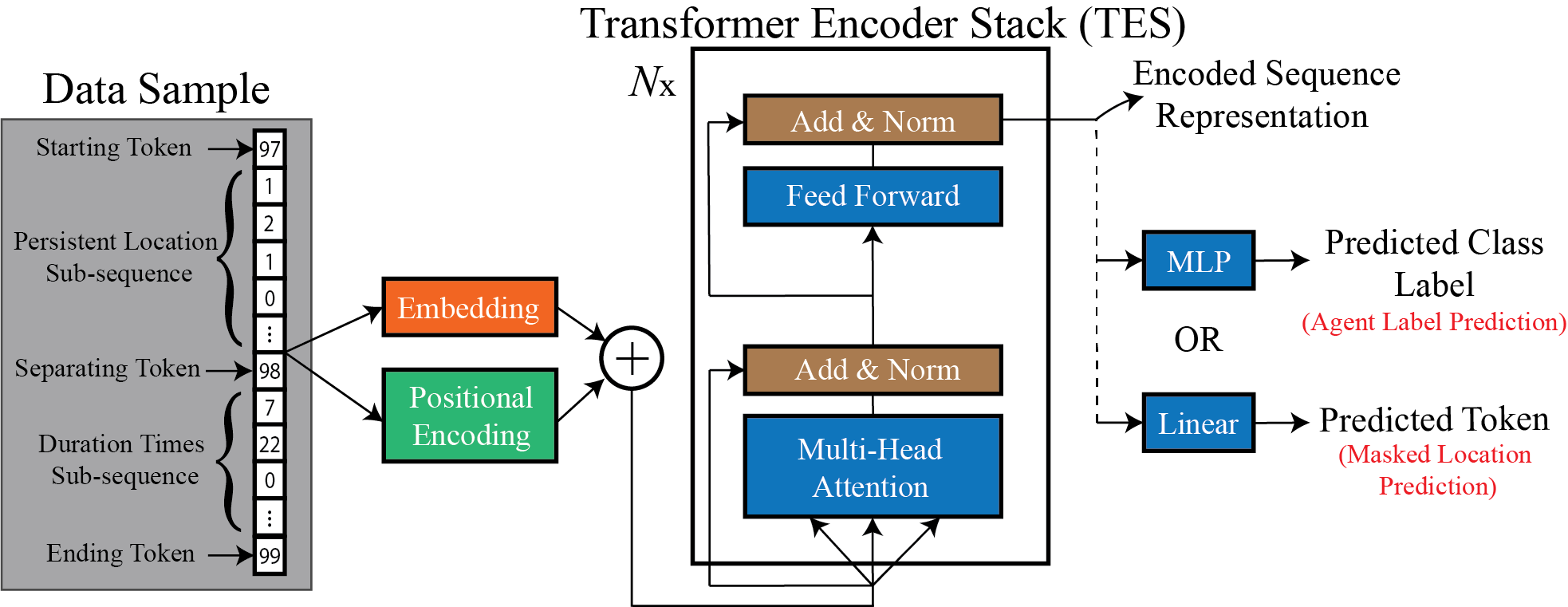}
    \caption{Our proposed STARE Transformer-based architecture, which uses a TES to create meaningful encodings of input data, an MLP to make classification predictions, and a Linear decoder to decode the encodings.}
    \label{fig:model}
\end{figure}
Figure \ref{fig:model} displays our model architecture, which takes in a sequence of tokenized data that consists of two sub-sequences, one with visited PLs, and the other with the duration times between the visited locations, and passes it through to a TES. The TES creates encodings, which it passes to a Multi-Layer Perceptron (MLP) to make classification decisions or a linear layer to decode the location encodings. The encodings from the TES can be leveraged to either learn labels of interest or understand relationships between data observations.  

\subsection{Benefits of STARE Model}

Through the use of either a classification or a masked location modeling task, our STARE model is able to learn relationships in data.  By learning agent labels through classification, we can learn \emph{agent} relationships. By learning tokens through a masked location token modeling task, we can learn relationships between \emph{locations}. To our knowledge, we are the first to apply the masked location modeling task to spatiotemporal data.

Our STARE model takes as input a data sequence, which is a concatenation of both location and time subsequences, and processes it; in contrast, these subsequences can be processing separately, like in the manner presented in \cite{Zhou:2022}, which uses separate spatial and temporal transformers.  In our approach, we choose to have a unifying model that concatenates spatial and temporal subsequences with a separating token, thus creating more synergies because all submodules in each encoder block (e.g. self-attention mechanisms, feedforward networks) will be a function of all the spatiotemporal information. Using two separate models can potentially break relationships in data or not discover them, specifically in places where they do not cross attend.

Additionally, our STARE model is designed to be a simple architecture; we have a flexible and general model that can be applied to any spatiotemporal data.  Our choice yields a few benefits:
\begin{itemize}
    \item We can apply our model to scenarios where road networks do not exist or are impractical (e.g. animal or other object mobility data). By not relying on road network information, we also prevent biases in our model from the road information being potentially out of date or inaccurate, whereas the road network intrinsically learned by our model will be correct for the time when the data was collected. 
    \item We have the flexibility to pretrain our model on massive amounts of trajectory data and then fine-tune it for specific tasks such as classifying agents in particular areas of interest (similar to how large language models are trained). 
    \item Our model is relatively easy to implement and does not require significant additional engineering to “make it work” for slightly different data scenarios, thus allowing for greater reproducibility. It is also less likely to be fragile to new data as it relies on learning structures within the data instead of being engineered to have certain structures. 
\end{itemize}

\section{Simulation \& Real Data Experiments}

\subsection{Dataset Format}
\label{dataset_format}

For our experiments, we mainly focus on datasets containing rich PoL trajectory information for various entities (i.e., humans, animals).  We aim to extract relationships between these entities and/or locations they frequent in their PoLs, so we make the implicit assumption that there exists a set of relationships in the data that we seek to identify.  Due to this, we do not focus on standard trajectory datasets (i.e., taxi trajectory\footnote{https://www.kaggle.com/c/pkdd-15-predict-taxi-service-trajectory-i/data}\footnote{https://www.kaggle.com/datasets/arashnic/tdriver}, small-scale human pedestrian mobility trajectories \cite{Lerner:2007}) as there might not be any latent relationships in the observed trajectories of these datasets.    

All datasets we consider have spatiotemporal information, i.e. \emph{latitude, longitude,} and \emph{timestamp}. Our model is specifically constructed for this type of data as opposed to generic time series data which can have any types of features over time. This allows us to:

\begin{enumerate}
\item Tokenize the time series in a meaningful way that allows us to capture continuous valued latitude, longitude coordinates in a finite vocabulary based on the spatial nature of the data. Thus, when performing masked location modeling, we can learn the meaning of these spatial tokens with respect to others (e.g., token A and token B will be highly correlated if they are frequented by the same agents).
\item Reduce the sequence length of the time series by only keeping tokens representing PLs allowing for each sequence to represent a much longer window of time (e.g., 1 day).
\end{enumerate}

\subsection{Metrics and Interpretations}
\label{metrics}

In every dataset that we consider, we have spatiotemporal trajectories that are labeled as belonging to a specific entity (i.e., an agent, animal).  Using these labels, we can carry out a label prediction task, and we can also do masked location prediction by masking random location tokens in our input sequences.  With either training regimen, we can obtain standard performance metrics (i.e., accuracy, precision, recall, F1-score, etc.); but, in our setting, we mainly care about presenting relationships that are learned in the data (i.e., between entities and locations), so we will only present accuracy metrics when discussing performance metrics.  We showcase our primary focus on the relationships that are learned by our model by presenting a slew of informative visuals and interpretations that highlight the interesting relations that our STARE model learns.      

\subsection{Simulated Data Experiments}
\label{subsection_experiments}

We simulate trajectories with a generative model that approximately aligns with our modeling assumptions for the STARE model. Specifically, our generative model is parameterized by a series of PLs associated with each agent. We also make the following assumptions in order to have realistic behaviors for the \emph{human} agents: They often: $1$) have reoccurring behaviors in the same locations (e.g., go to the same house every day), $2$) live and behave similarly to others (i.e., there is some sharing of PLs between humans), and $3$) travel along roads between their PLs.

Thus, the process follows a hidden Markov model (HMM) backbone where the states are specific locations extracted from various foundational data sources \cite{usastructures, planetsense, OSM} and fall into the following categories: home, work, food, and gym locations. Each agent has its own random transition matrix with non-zero entries for a subset of the states (e.g., each agent only has one unique state it can transition to for its home and work locations). Additionally, in order to encourage more realistic group and social behaviors, agents' transition matrices are generated hierarchically with a grouping parameter such that agents that share the same group (i.e., belong to the same subpopulation) are assigned houses nearby each other or share the same office location. Finally, in order to create high frequency, realistic trajectories, we interpolate between the states using an Open Street Maps (OSM) road network from the OSMnx python package, \cite{osm_python}. Two examples of multiple generated trajectories that belong to the same subpopulation are shown in Appendix \ref{app:vis_subpop}.  In general, this simulation approach allows for the creation of balanced classes of subpopulations.

For our experiments in this subsection, we simulated data trajectories with $10$, $20$, $30$, and $50$ subpopulations over a $28$ day time period with $37$, $348$, $1,288$, and $10,909$ agents, respectively. Given these datasets, we train our STARE model with a $4$ attention head, $2$ encoder stack architecture and use an $85$\%/$15$\% train/test split. We tokenize each agent's trajectories into a $1$ day time window, map the stationary points to an alphabet based on Zoom $16$ S2 cells, and map the duration times to an alphabet of $10$ minute increments. Different choices of S2 cell sizes and duration time increments will alter the model vocabulary size and the resolution of the discretized data.

\subsubsection{Classification Results}
Since we know which agents belong to which subpopulations, we train our STARE model to learn either subpopulation or agent labels. In a realistic setting where subpopulation labels are not available, we aim to use agent labels as a learning task that yields informative embeddings that can be leveraged to create a notion of subpopulations.  

\begin{table}[H]
  \centering
  \begin{minipage}{\linewidth}
    \centering
    \captionsetup{justification=centering} % Center the subcaption
    \caption*{Agent Label Classification Accuracy} % Subtitle for Table 1
    \vspace{-10pt} % Adjust vertical space between subtitle and table
    \begin{tabular}{|c||c|c|c|c|}
    \hline
        Dataset & STARE & \multicolumn{1}{c|}{LSTM} & BiLSTM \\
        \hline 
        (S) & 94.2\% & 88.7\% & 89.9\% \\
        \hline
        (M) & 78.6\% & 77.5\%  & 78.4\% \\
        \hline
        (L) & 73.1\%  & 73.9\% & 74.1\% \\
        \hline
        (XL) & 51.7\% & 52.7\% & 53.8\% \\
        \hline
    \end{tabular}
  \end{minipage}
\vspace{20pt}% 
  \begin{minipage}{\linewidth}
    \centering
    \captionsetup{justification=centering} % Center the subcaption
    \caption*{Subpopulation Label Classification Accuracy} % Subtitle for Table 2
    \vspace{-10pt} % Adjust vertical space between subtitle and table
    \begin{tabular}{|c||c|c|c|c|}
    \hline
        Dataset & STARE & \multicolumn{1}{c|}{LSTM} & BiLSTM \\
        \hline 
        (S) & 100\% & 100\% & 100\% \\
        \hline
        (M) & 100\% & 100\%  & 100\% \\
        \hline
        (L) & 100\%  & 99.988\% & 100\% \\
        \hline
        (XL) & 99.993\% & 99.957\% & 99.989\% \\
        \hline
    \end{tabular}
  \end{minipage}
  \caption{Accuracy of our STARE model, along with baseline LSTM and BiLSTM architectures, on simulated datasets of various sizes and labeling schemes. The (S)/(M)/(L)/(XL) datasets have $10/20/30/50$  subpopulations with $37/348/1288/10909$ agents total, respectively. Our STARE model matches and sometimes exceeds the baselines' accuracies.}
  \label{tab:accs}
\end{table}

In Table \ref{tab:accs}, we show the accuracy (i.e., correct classification rate) for each training approach on our $4$ simulated datasets. Additionally, we compare our method against LSTM and BiLSTM baselines, of the same amount of stacks as the amount of encoders in our STARE model, with an Adam optimizer and embedding and hidden dimensions of sizes $128$ and $64$, respectively; we see that all models tend to yield similar accuracies. When training on subpopulation labels, STARE has essentially perfect accuracy, but when training on agent labels, there are some misclassifications due to similar PoLs between agents in the same subpopulation. 

\begin{figure}[H]
\centering
\includegraphics[width=0.9\linewidth]{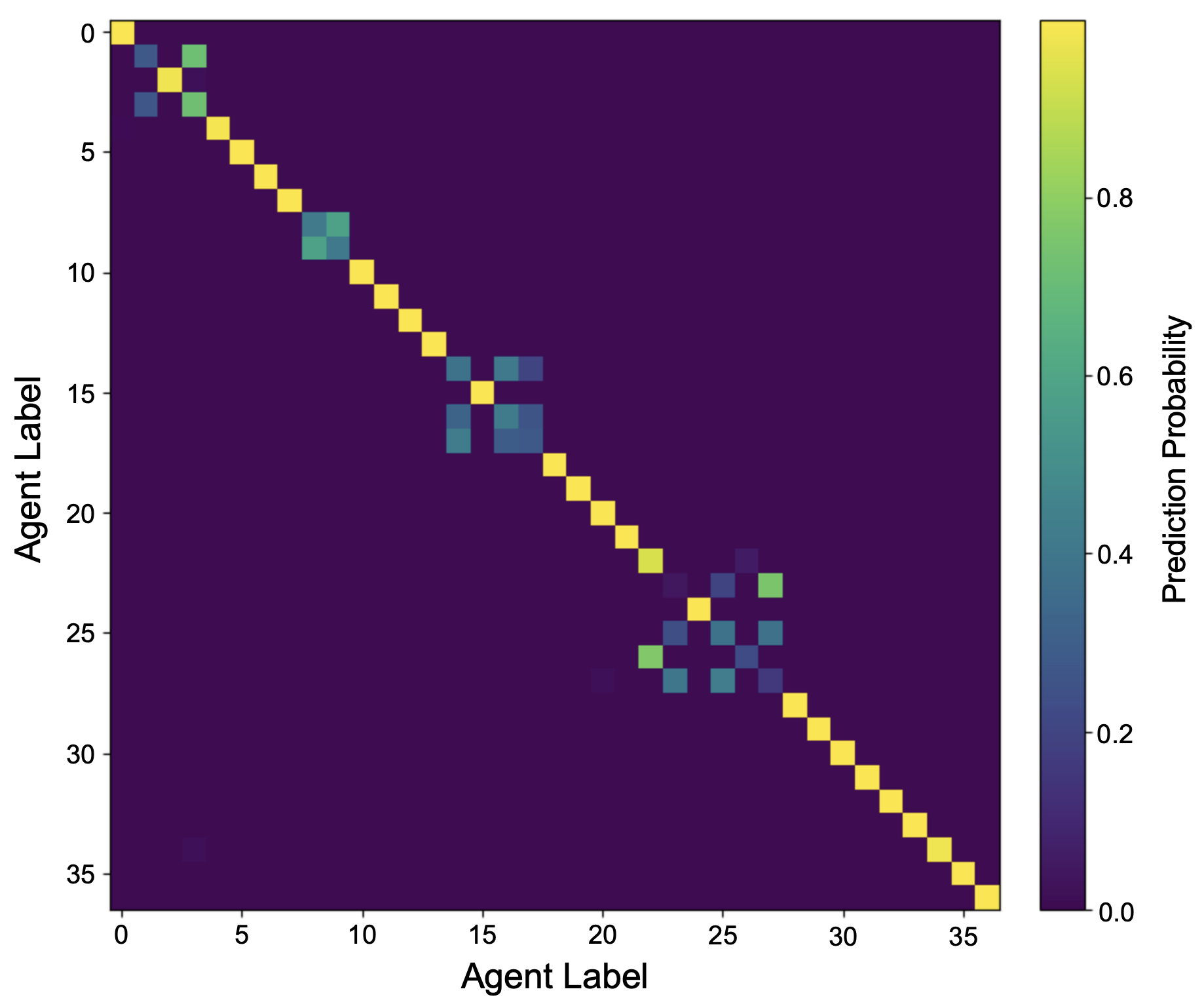}
\caption{Matrix of average predicted probability scores for the (S) dataset (consists of $37$ agents) where the rows and columns are the true and predicted agent labels, respectively.}
\label{fig:conf_labels}
\end{figure} 

\begin{figure}[H]
    \centering
    \includegraphics[width=\linewidth]{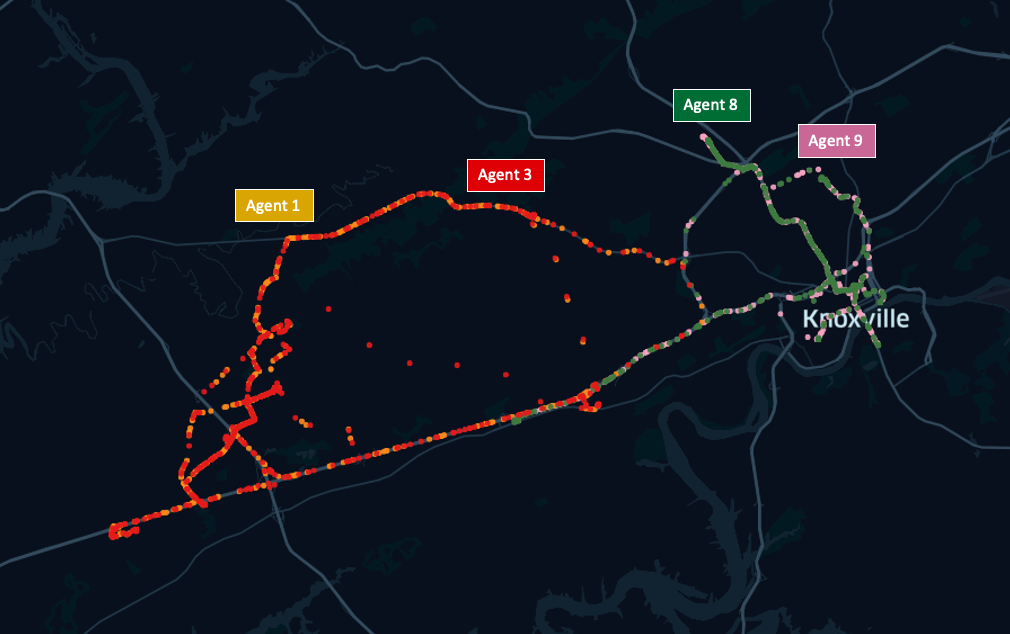}
    \caption{Visualization of the similar PoLs between misclassified agents. The agents corresponding to the $1$st/$3$rd rows of the matrix shown in Figure \ref{fig:conf_labels} are in red/yellow and those for $8$th/$9$th rows are in green/pink.}
\label{fig:example_misclassifications}
\end{figure} 

Figure \ref{fig:conf_labels} represents a matrix of averaged predicted probability scores for the (S) dataset obtained from training the STARE model on agent labels; every row represents an agent class. We see that many agents are correctly identified (only non-zero value is in the diagonal), but there are others that have multiple blocks of non-zero values. Figure \ref{fig:example_misclassifications} visualizes the trajectories of the agents in the first two blocks where we can see the reason for the misclassification; the PoLs between these agents are extremely similar. With this, we see that we can leverage misclassifications of our STARE model to learn relationshiops between agents.

To ensure informative embeddings, our model should learn something from agent labels, so the accuracies we obtain should be relatively high, as they indeed are. If not, the obtained embeddings may not be informative as the model does not learn well. In many realistic scenarios, subpopulation information will not be present, so we can use agent labels to learn informative encodings from the output of the TES and leverage them to learn relationships between agents in an attempt to form subpopulations. 

\begin{figure}[H]
    \centering
    \includegraphics[width=.975\linewidth]{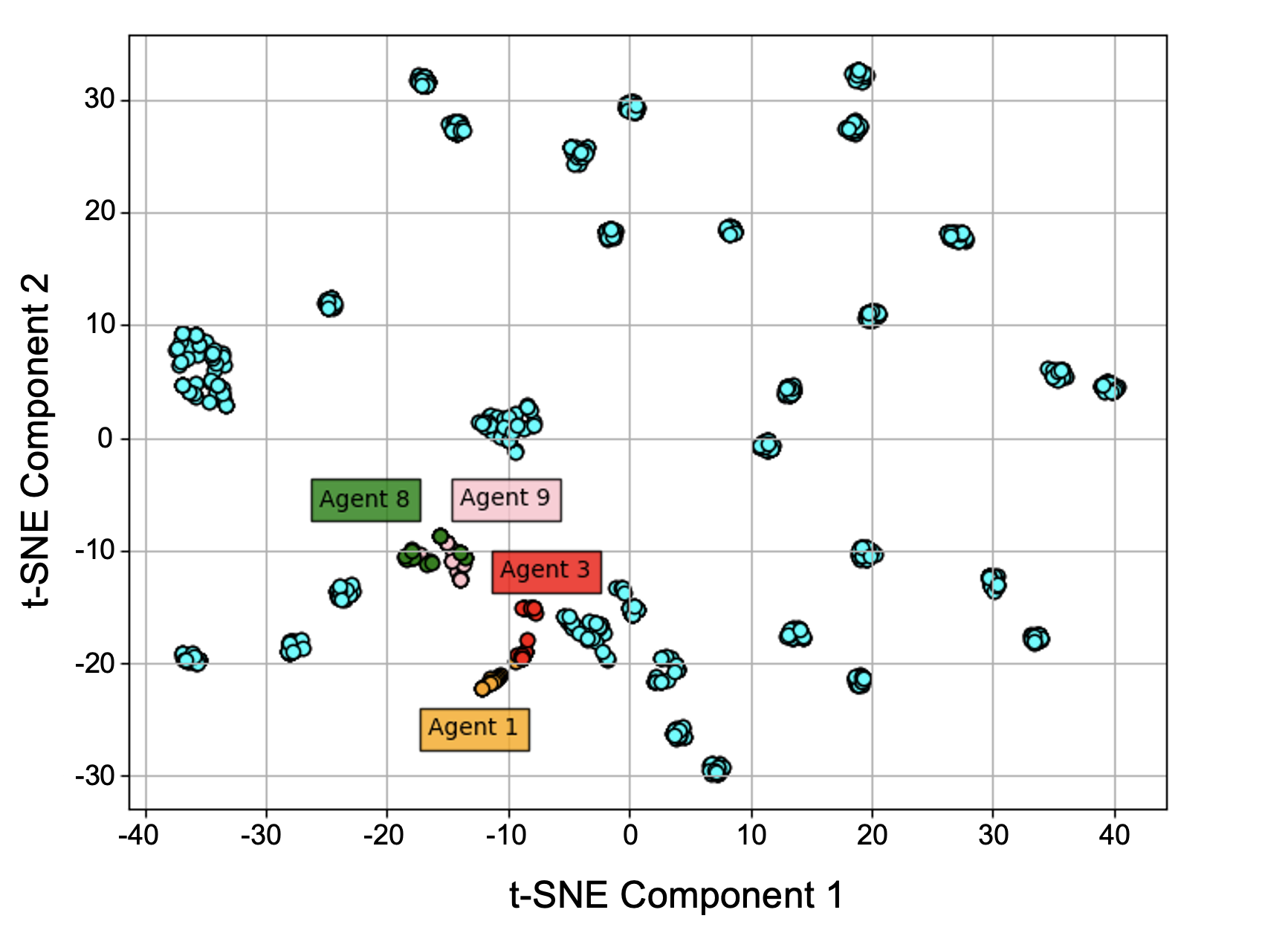}
    \caption{Low dimension embeddings obtained from applying t-SNE on the high dimensional TES embeddings for the (S) dataset. We see that agents that belong to the same underlying subpopulations tend to lie close to each other in the embedding space, as can be seen with agents [$1$,$3$] and agents [$8$,$9$], which have consistent colors with those in Figure \ref{fig:example_misclassifications}.}
\label{fig:tsne_MB}
\end{figure}

Next, we take the data encodings from the output of the TES when training on agent labels and apply t-SNE to get a two dimensional representation of the embedding space, as shown in Figure \ref{fig:tsne_MB}. Here, we see that there are many distinct clusters of data points that represent specific agents.  Some agents are similar though, so their data points are proximal in the embedding space, as can be seen with the two pairs of agents [$1$,$3$] and [$8$,$9$] in Figure \ref{fig:tsne_MB}.  With this, one can apply a clustering method in this lower dimensional embedding space to group similar agents together and, in turn, form subpopulations and learn agent relationships.  

\subsubsection{Masked Modelling Results}

In additional to the classification task, we can also train our model with a masked location modelling task on the same datasets as the classification experiments to see how STARE performs at learning the distribution of the sequences.

\begin{figure}[H]
\centering
\includegraphics[width=0.785\linewidth]{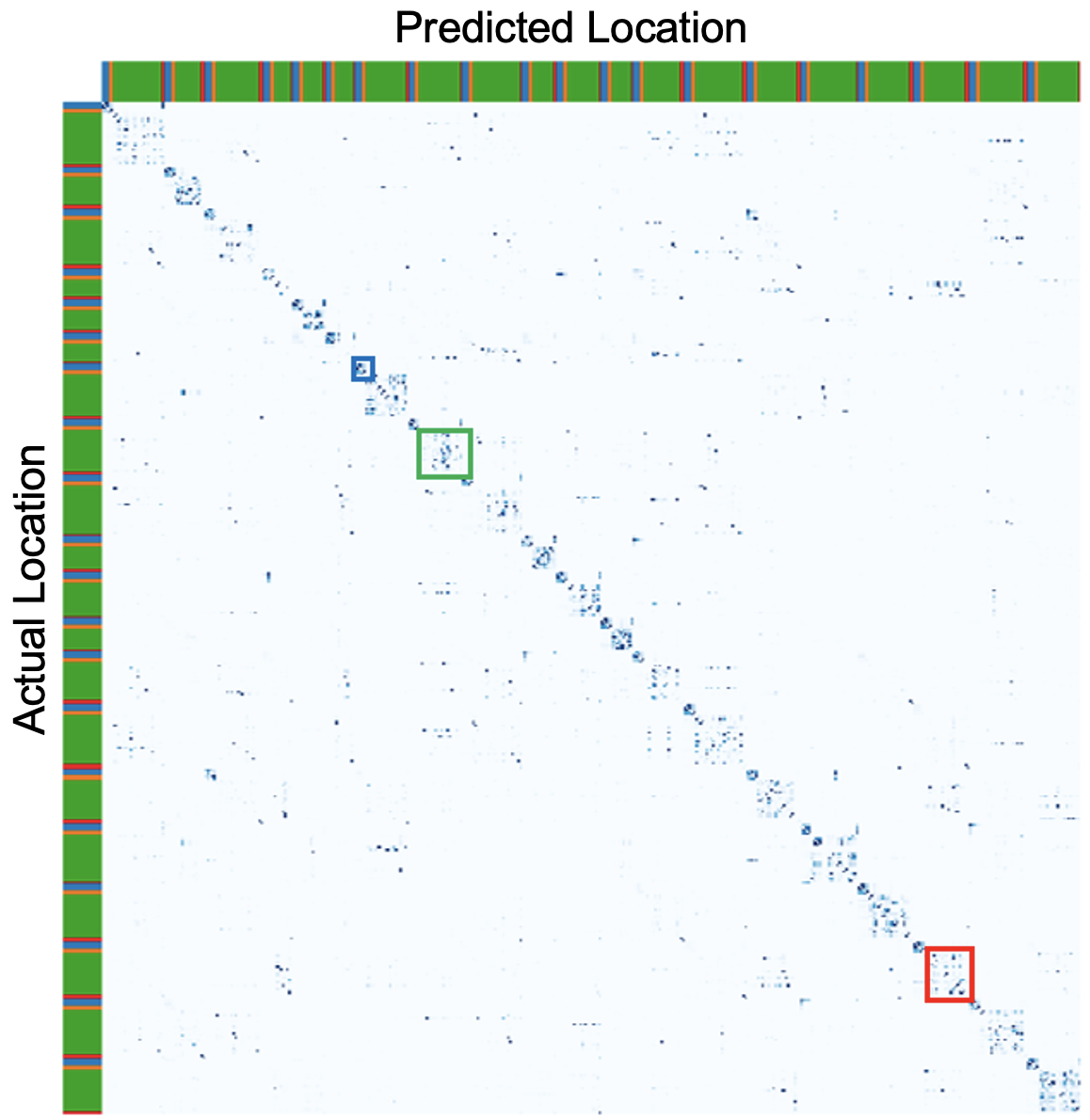}
\caption{Matrix of average predicted probability scores for the (M) dataset where the rows are the true $342$ S2 cell locations and the columns are the predicted locations. In the colored bands along each axis, the blue, orange, green, and red colors represent food, gym, home, and workplace locations, respectively.}
\label{fig:masked_conf_labels}
\end{figure}

In Table \ref{tab:acc_masked1}, we show the accuracies of our STARE model at masked location modelling. Masked location modelling (or masked language modelling in NLP) is a unique feature of transformer architectures and thus we do not have any comparisons to LSTM-style architectures. However, we can analyze the misclassifications in a similar fashion as those of the preceding label classification experiments were leveraged to learn interesting agent relationships. In this case, misclassifications indicate similarity between locations. Figure \ref{fig:masked_conf_labels} shows a matrix of the averaged predicted probability scores for each S2 cell location for the (M) dataset obtained from training the STARE model on agent labels.  

\begin{table}[H]
\centering
\begin{tabular}{|c||c|c|c|c|}
\hline
      & (S) & (M) & (L) & (XL) \\
\hline
STARE & 85\% & 72.3\% & 76.9\% & 77.8\% \\
\hline
\end{tabular}
\caption{Masked location modeling accuracy performance table of our STARE model; we see that STARE performs well at learning the marginal distributions of the sequences.}
\label{tab:acc_masked1}
\end{table}

We can permute the rows and columns of this probability matrix roughly based on subpopulation information, where as described in the data generating process, agents that share the same subpopulation are assigned houses nearby each other, or they often share the same office location. Here, we see a block structure characterized by misclassified locations and governed by both the specific type of location and the subpopulation membership.  As with the agent label training approach, we can study these misclassifications to understand relationships between locations. We focus on three misclassification blocks in Figure \ref{fig:masked_conf_labels} and we highlight them using blue, green, and red rectangles. The green and red rectangles correspond to misclassified home locations, while the blue rectangle corresponds to misclassified food locations. We now look at the PoLs of the agents that reside in their respective misclassification blocks.  %The following three figures present the PoLs of the agents that reside in their respective misclassification blocks.

\begin{figure}[H]
    \centering
    \includegraphics[width=0.9\linewidth]{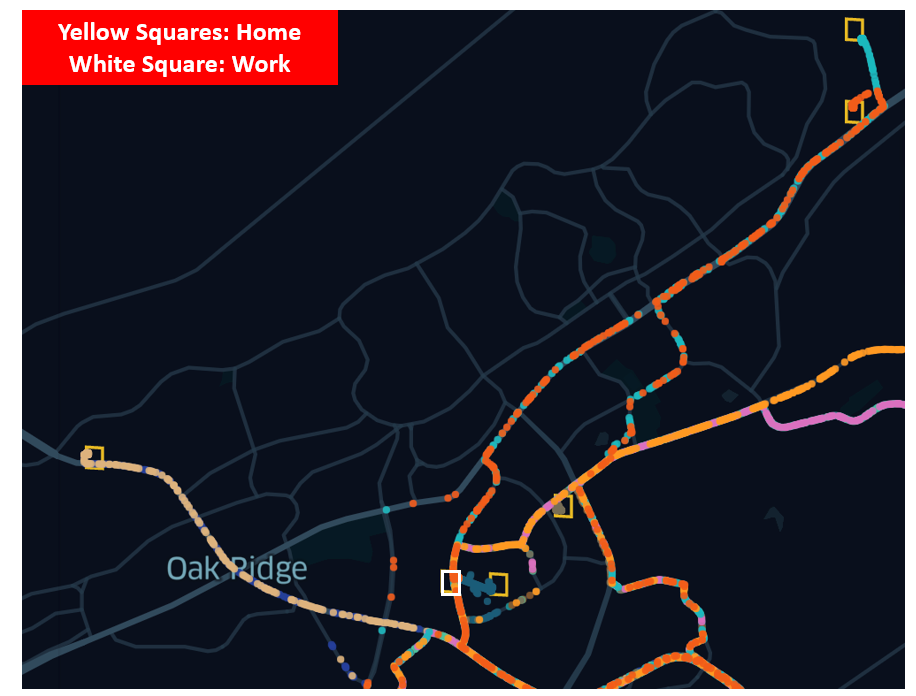}
    \caption{Visualization of the PoLs of agents residing in the green square in Figure \ref{fig:masked_conf_labels} which shows similarly predicted S2 cells containing homes of agents in the same subpopulation.}
\label{fig:masked_green_visual}
\end{figure}

Figure \ref{fig:masked_green_visual} focuses on the green rectangle where home locations get misclassified.  Specifically, we extract a subset of agents that have PLs that overlap with the misclassified ones in the green rectangle and we first visualize their PoLs.  We see that these agents all share the same work location (shown in the white square), but do not share the same home locations (shown in the yellow squares).  Overall, there are similar agents that differ based on their homes, so, in turn, these home locations understandably get misclassified.  From this, we see that these misclassifications can be leveraged to showcase interesting connections between agents and locations.  Interpretations of the results from the blue and red rectangles are presented in Appendix \ref{app:sim_1}.

\subsection{``Massive Trajectory Data Based on Patterns of Life" Experiments}

We also apply our STARE model to a more complex simulated dataset from \cite{amiri2023massive}, which contains movement data for $10,000$ agents in 3 metropolitan areas (Atlanta, New Orleans, and George Mason University) for 15 months. These experiments allow us to show the generalizability of STARE to other datasets where the generating assumptions do not perfectly align with our modeling assumptions and to showcase STARE's ability to scale to large-scale, long duration datasets. 

Interestingly, although the data sampling rate of this dataset is every $5$ minutes, the agents tend to move in a \emph{jumpy} manner. However, because our STARE model only needs stationary points and the times spent at these points, we can still tokenize each agent’s trajectories into daily sequences in a similar fashion as in the previous experiment. We drop the first month of data as suggested by \cite{amiri2023massive}, who indicates that it contains non-converged "warm-up" data from their simulator. 

We trained our STARE model using agent labels with a larger architecture (4 encoder stacks, instead of 2) and the same train/test split percentages used in the previous set of experiments. These models took roughly 14 to 16 hours to train for 2000 epochs and their test accuracies are shown in the first row of Table \ref{tab:accs2}. We additionally compared against similarly sized LSTM/BiLSTM baseline models.

% LSTM, BiLSTM, 55.8
\begin{table}[H]
\centering
\begin{tabular}{|c||c|c|c|c|}
\hline
      & ATL & NOLA & GMU \\
\hline
STARE & 88.1\% & 89.1\% & 79.8\% \\
\hline
LSTM & 66.4\% & 65.9\% & 51.6\% \\
\hline
BiLSTM & 73.5\% & 73.1\% & 62.6\% \\
\hline
\end{tabular}
\caption{Accuracy of our STARE model, along with baseline LSTM and BiLSTM architectures, on classifying agent labels for the Atlanta (ATL), New Orleans (NOLA), and George Mason University (GMU) datasets.  Our STARE model yields larger accuracies than the baseline models when training on agent labels.}
\label{tab:accs2}
\end{table}

\begin{figure}[H]
    \centering
    \includegraphics[width=\linewidth]{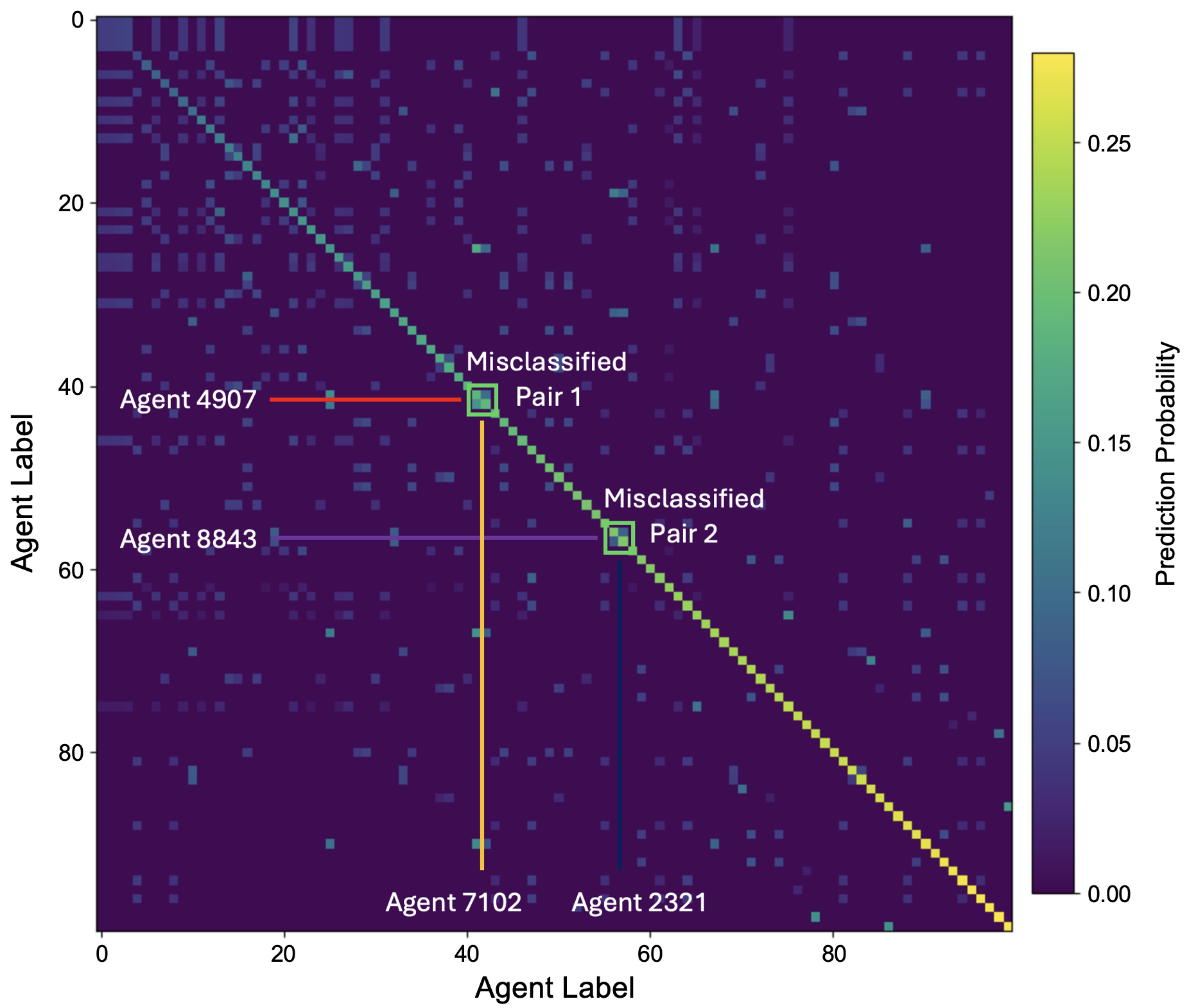}
    \caption{Averaged post-softmax prediction matrix of the STARE model for the top $100$ misclassified agents.  These misclassifications can be leveraged to learn relationships between agents. }
    \label{fig:atlanta_matrix}
\end{figure}

\begin{figure}[H]
    \centering
    \includegraphics[width=\linewidth]{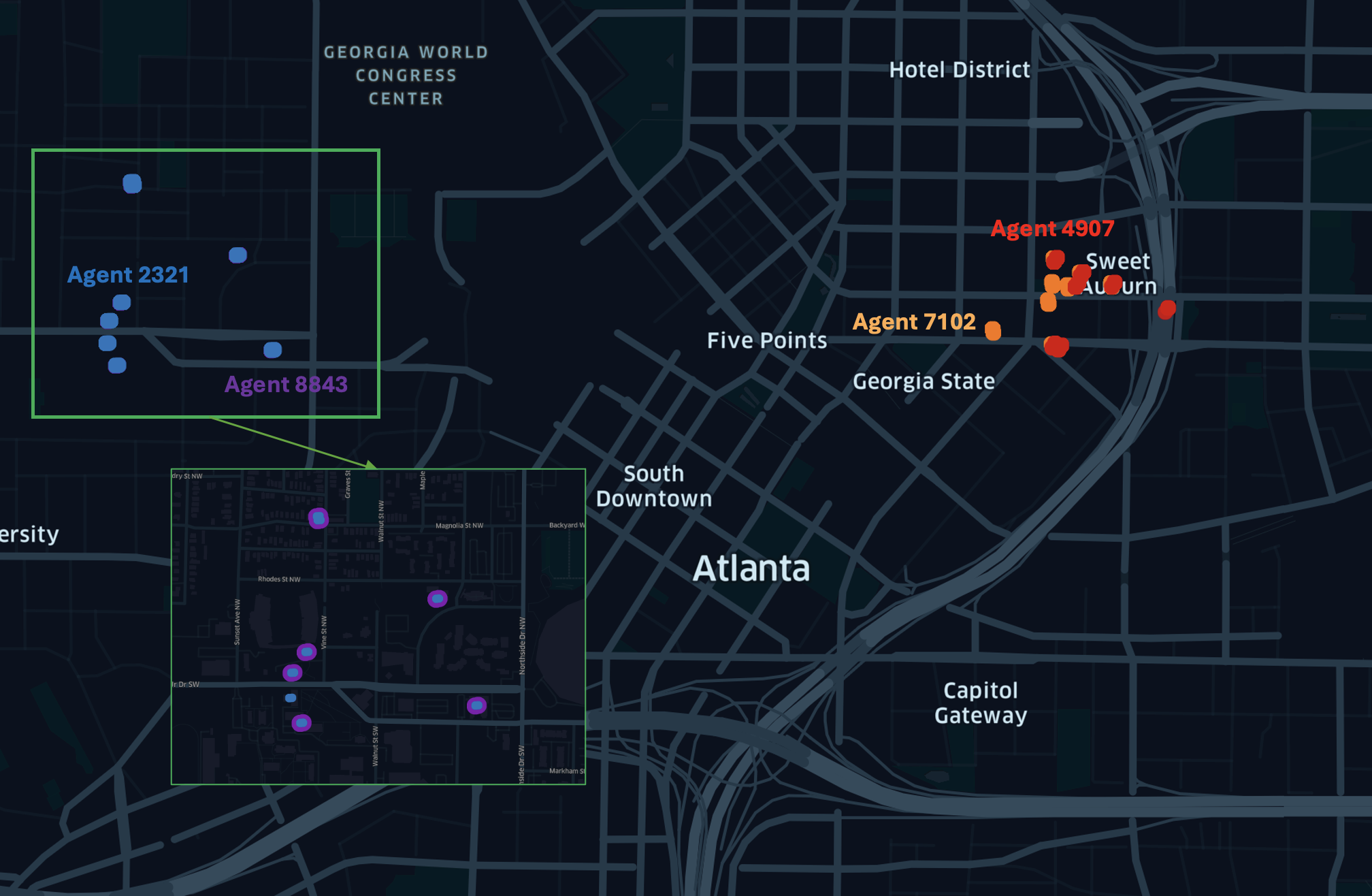}
    \caption{PLs for the 2 pairs of misclassified agents: [$7102$, $4907$] and [$2321$, $8843$]. We see that the agents in each pair have similar PoLs; those for agents $2321$ and $8843$ are extremely similar and even lie on each other.}
    \label{fig:atlanta_s2}
\end{figure}

\begin{figure}[H]
  \centering % Centers the entire figure
    \includegraphics[width=0.95\linewidth]{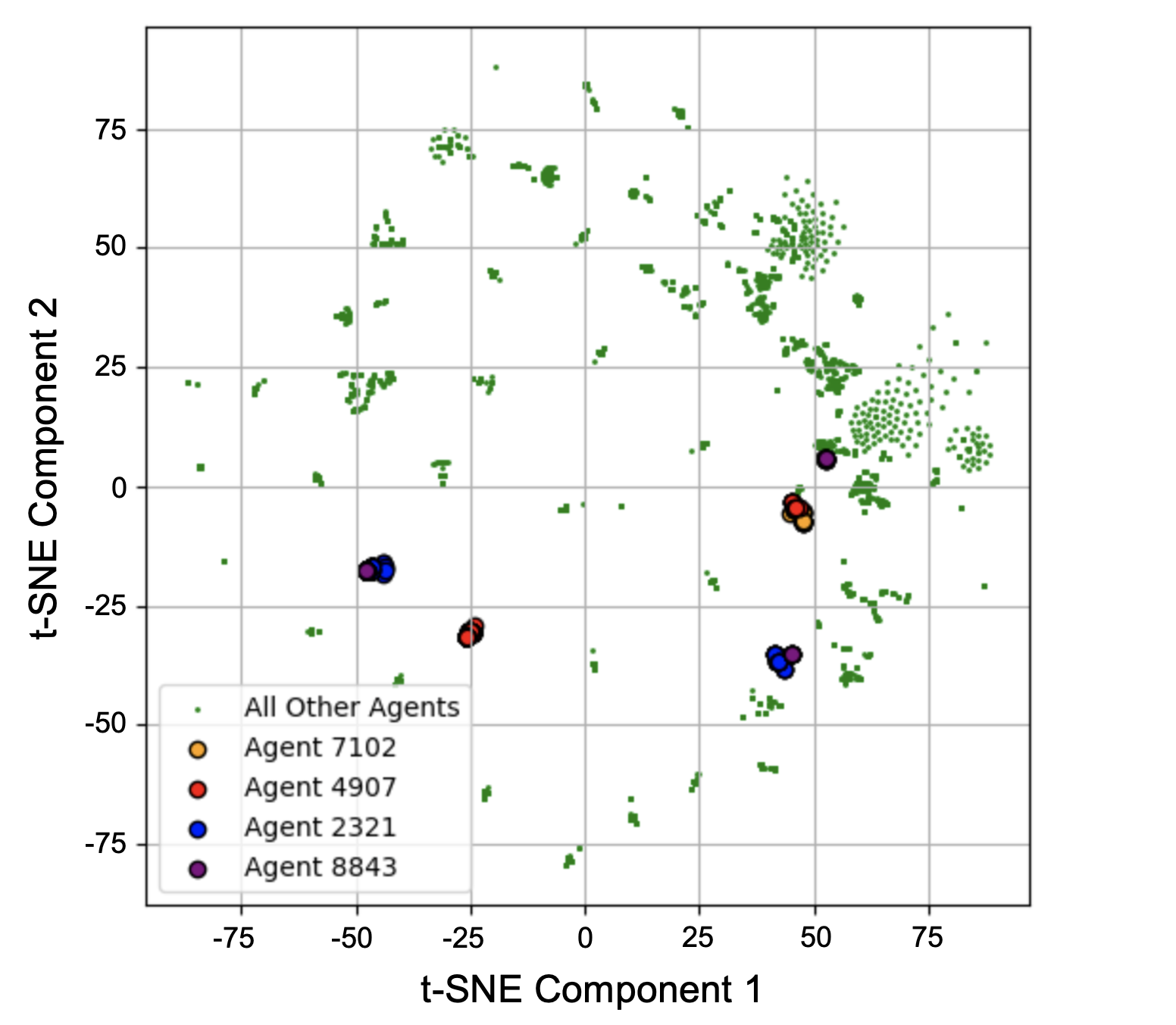}%
    \caption{Low dimension embeddings obtained from applying t-SNE on the TES embedding for the ATL dataset. We see that agents that are similar in their PoLs tend to lie close to each other in the embedding space, as can be seen with the misclassified agent pairs: [$7102$, $4907$] and [$2321$, $8843$].}
    \label{fig:ATL_tsne}
\end{figure}

We also visualize the top $100$ misclassified classes from the predicted post-softmax probability matrix in Figure \ref{fig:atlanta_matrix} and choose $2$ highly misclassified pairs of agents: [$7102$, $4907$] and [$2321$, $8843$], to study further. In Figure \ref{fig:atlanta_s2}, we see that these agent pairs have very proximal PLs and many of their locations are tokenized into the same S2 cells.  In turn, the PoLs of these agent pairs are very similar and end up getting misclassified by the STARE model. We apply t-SNE to the TES data encodings and visualize the low-dimensional embedding space in Figure \ref{fig:ATL_tsne}.  Specifically, we look at the locations of data observations for the $2$ agent pairs of interest, and we see that the pairs are close in the embedding space, as expected. Ultimately, like with the previous experiment, we can learn relationships between agents by studying model misclassifications and we can even apply a clustering method to the embedding space to further learn these relationships.  

\newpage

\subsubsection{Masked Modelling Results}

We showcase the accuracy of the STARE model on the masked location modelling task in Table \ref{tab:acc_masked2}.  We additionally present further results on interpreting the output of this task in Appendix \ref{app:sim_2}.  

\begin{table}[H]
\centering
\begin{tabular}{|c||c|c|c|c|}
\hline
      & ATL & NOLA & GMU \\
\hline
STARE & 77.7\% & 79.58\% & 78.51\% \\
\hline
\end{tabular}
\caption{STARE accuracies from the masked location modelling task on the ATL, NOLA, and GMU datasets. }
\label{tab:acc_masked2}
\end{table}

\subsection{Wildlife Animal Movement Data Experiments} 

To assess the effectiveness of extracting meaningful encodings with STARE in real scenarios, we applied it to a raw trajectory dataset of ravens \cite{ravens}. We only used trajectory data from $49$ ravens between March and July for both $2018$ and $2019$ because we observed a higher volume of data for these intervals (the GPS sensors on the ravens are solar and data observations are very infrequent during the winter months). We tokenized each agent’s trajectories into $1$ day time window and altered the S2 Cell Zoom Level from 16 to 14. Otherwise, we trained our STARE model on the raven dataset using the same model architecture and hyperparameters as in the first set (Section \ref{subsection_experiments}) of simulated data experiments. We also train baseline LSTM/BiLSTM models using agent labels with the same training settings as before and obtain the accuracies shown in Table \ref{tab:accs3}. 

\begin{table}[H]
\centering
\begin{tabular}{|c||c|c|c|c|}
\hline
      & STARE & LSTM & BiLSTM \\
\hline
Ravens & 39.5\% & 34.6\% & 37.8\% \\
\hline
\end{tabular}
\caption{Accuracy of STARE compared against the baseline LSTM and BiLSTM models on the ravens dataset when training using agent labels.}
\label{tab:accs3}
\end{table}

In comparison to the accuracies seen in the preceding datasets, our tested models yield low accuracies.  We believe this happens due to a characteristic of the ravens dataset. As we discover in the next figures, there are $4$ main classes of ravens, and distinguishing between the classes is trivial.  The more difficult task is to distinguish between individual ravens in each class, since each raven acts in a similar manner (i.e., have similar PLs and duration times), perhaps due to the observed ravens moving in groups. Because of this difficulty, we observe smaller accuracies when training on individual raven labels. 

\begin{figure}[H]
  \centering % Centers the entire figure
    \includegraphics[width=0.95\linewidth]{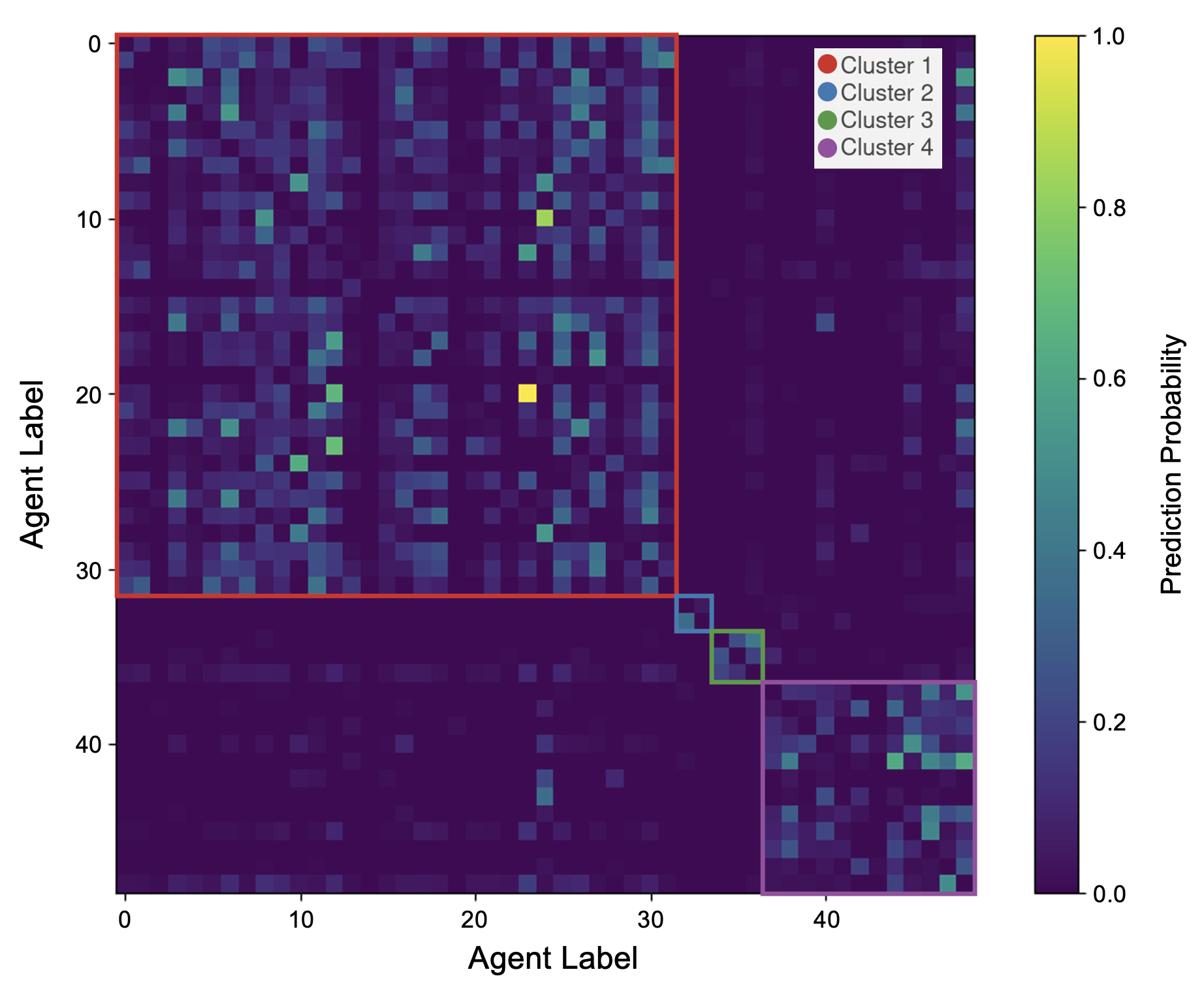}%
    \caption{The averaged predicted probability matrix after being clustered into $4$ clusters. For visualization purposes, the matrix is normalized by its maximum off-diagonal entry and the diagonal values are dropped.}
    \label{fig:raven_spectral_clusters}
\end{figure} 

\begin{figure}[H]
  \centering % Centers the entire figure
    \includegraphics[width=\linewidth]{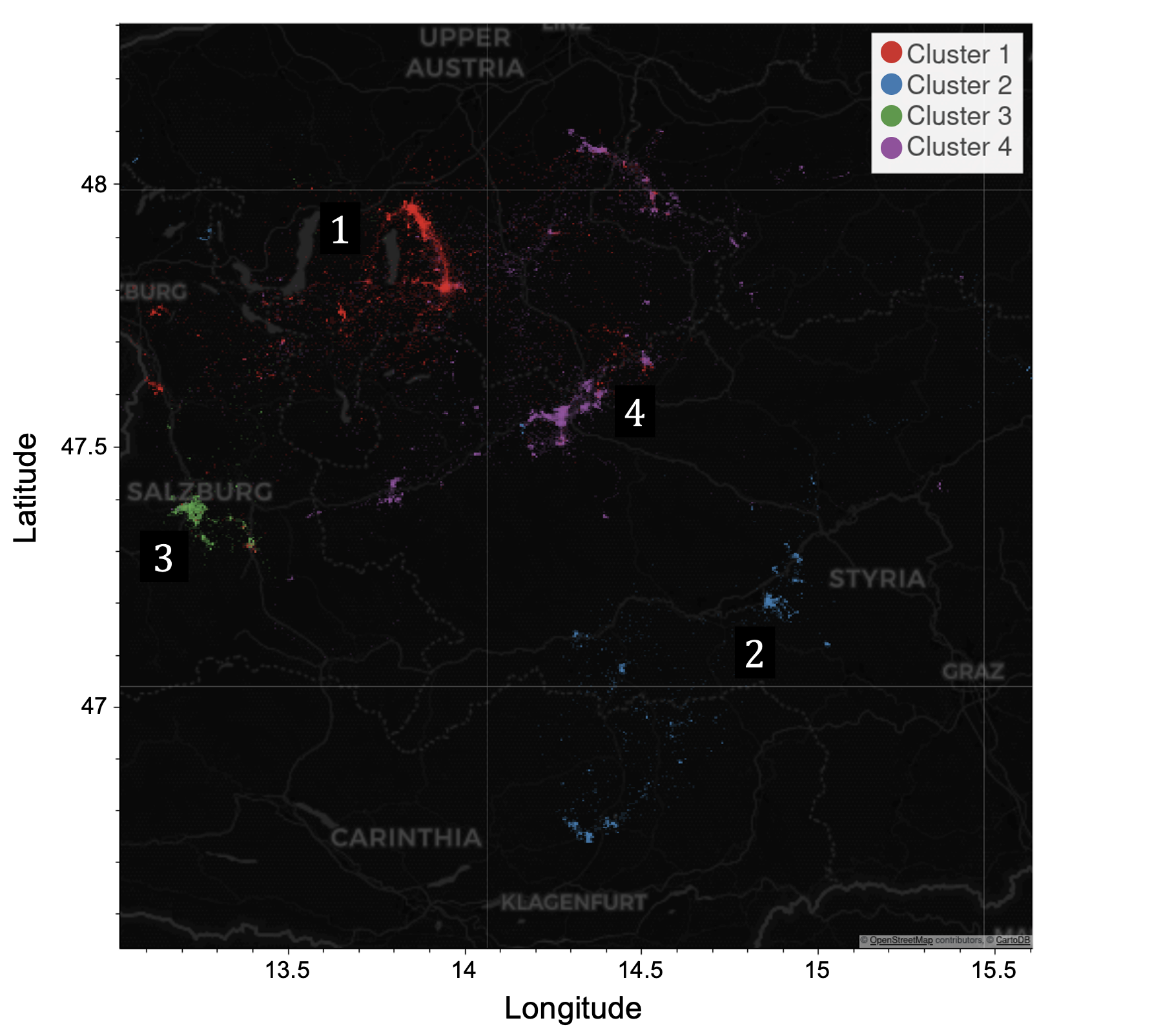}%
    \caption{All raven trajectories color coded by the cluster membership defined in Figure \ref{fig:raven_spectral_clusters} (intensity of a point is dependent on frequency of visits to the specific location). The clusters obtained from SC are spatially separated.}
    \label{fig:raven_clusters_traj}
\end{figure}

To further explore the potential of clustering raw spatiotemporal data using the TES encodings, we apply an MLP to them and obtain an averaged post-softmax prediction score matrix; in turn, we apply Spectral Clustering (SC) to it. Figure \ref{fig:raven_spectral_clusters} shows the results of SC, where $4$ distinct clusters are identified along the diagonal from top to bottom, despite the second cluster containing only two agents. Figure \ref{fig:raven_clusters_traj} also displays all $49$ ravens color-coded by cluster membership. This visualization reveals clear distinctions between the 4 groups, demonstrating the efficacy of our approach in obtaining meaningful embeddings that can be leveraged to understand spatiotemporal data patterns and similarity between agents. Additionally, we apply t-SNE to the TES data encodings and showcase a visual of the low-dimensional embedding space in Appendix \ref{app:ravens}. %Figure \ref{fig:raven_tsne}. All the data points are colored based on the cluster membership defined in Figure \ref{fig:raven_spectral_clusters}; we see that the $4$ clusters obtained from SC are also separated in the embedding space, thus further giving us confidence that our STARE model is able to obtain meaningful embeddings that can be used to learn relationships between agents.

We do not perform masked location modelling on the ravens dataset because of the lack of semantic spatial meaning in the locations frequented by ravens along with the much lower resolution of the S2 cells (zoom 14) used due to much higher noise level of the GPS sensors on the ravens.

\section{Conclusions and Future Work}

In this work, we proposed a spatiotemporal transformer-based model for extracting relationships from PoL data along with meaningful data encodings. Through experiments, we demonstrated our model's ability to correctly learn embeddings from two different tasks: agent label classification and masked modelling. Additionally, we presented various ways of understanding relationships between similar agents and locations through the use of the obtained embeddings. As future work, we plan to investigate our model's ability to scale up to very large volumes (millions) of data and also develop a means of jointly carrying out masked language modeling and label classification to further extract interesting information from rich spatiotemporal PoL data.

% In this paper, we make the following contributions.  First, we present a data discretization technique for reducing the dimensionality of long and rich agent PoL data.  Second, we propose a novel transformer-based architecture for obtaining informative data embeddings which can be used for a variety of tasks.  Finally, we present extensive experiments on both simulated and real trajectory datasets and showcase our proposed STARE model's informative embeddings along with improved performance over baseline sequence encoder models (i.e., LSTM, BiLSTM) in regards to classification accuracy.  %(which can blow up in size given a high data sampling rate and large timeframe for data collection) 

\begin{acks}
Supported by the Intelligence Advanced Research Projects Activity (IARPA) via Department of Interior/ Interior Business Center (DOI/IBC) contract number 140D0423C0046. The U.S. Government is authorized to reproduce and distribute reprints for Governmental purposes notwithstanding any copyright annotation thereon. Disclaimer: The views and conclusions contained herein are those of the authors and should not be interpreted as necessarily representing the official policies or endorsements, either expressed or implied, of IARPA, DOI/IBC, or the U.S. Government.
\end{acks}


\bibliographystyle{ACM-Reference-Format}
\bibliography{mybibfile}

\clearpage
\appendix
\section{Appendix}
\renewcommand{\thefigure}{A\arabic{figure}}

\subsection{Data Discretization Example} \label{subsec:example_discretization}

\begin{figure}[!h]
    \centering
    \includegraphics[width=0.905\linewidth]{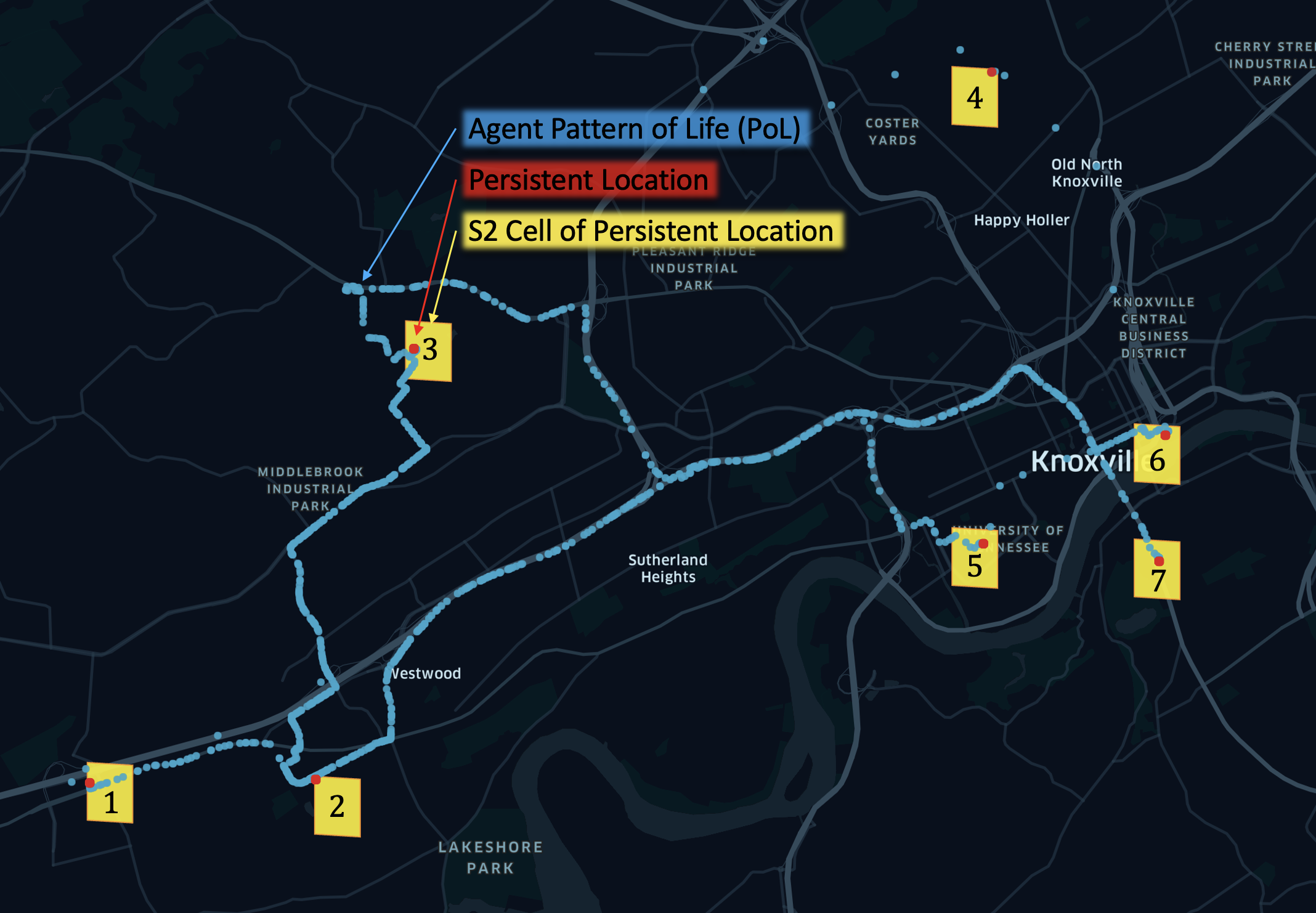}
    \caption{Visualization of an agent's PoL along with its derived PLs and the S2 cells that each PL maps to.} 
    \label{fig:example}
\end{figure}

As a guiding example, Figure \ref{fig:example} displays an agent's entire PoL along with their PL information. The data in blue represents the PoL defined by their GPS positions, the red represents the PLs, and the yellow rectangles represent S2 cells.

To form a data point for agent $a$ in time window $m$, we must create visited PL and stay duration time sub-sequences.  We begin with $T_{a,m}$, which is the raw trajectory of agent $a$ in time window $m$. Let us say that in time window $m$, agent $a$ starts at $\text{PL}_1$, goes to $\text{PL}_2$, then to $\text{PL}_5$, then to $\text{PL}_3$, and finally ends the day back at $\text{PL}_1$. We map each PL to a S2 cell of a certain zoom level and then we define the S2 cells using indices from $1$ to the maximum amount of S2 cells, $n_{\text{S2}}$.  The location sequence for the day becomes: $s(T_{a,m}) = [1, 2, 5, 3, 1]$.  With this, we then look at the duration of time that the agent stays at each visited location and form a sequence of these times as: $dt(T_{a,m}) = [t_{\text{S2}_1}, t_{\text{S2}_2}, t_{\text{S2}_5}, t_{\text{S2}_3}, t_{\text{S2}_1}]$.  Realistically, this sequence could be: $dt(T_{a,m}) = [8\text{hrs}, 6\text{hrs}, 2\text{hrs}, 2\text{hrs}, 6\text{hrs}]$. The summation of all of the values of $dt(T_{a,m})$ will equal $24$ hours, which represents the amount of hours in a day. Next, we adjust $dt(T_{a,m})$ by discretizing time into discrete blocks, assigning integers to each block, and offsetting these indices based on the total amount of S2 cells with PLs in the data ($n_{\text{S2}})$, so as to avoid token overlap in the input between positions and times.

For example, lets assume that the blocks are $30$ minute intervals and $n_{\text{S2}} = 20$ so our time sequence is: $dt(T_{a,m}) = [16, 12, 4, 4, 12] + n_{\text{S2}} = [36, 32, 24, 24, 32]$, where each element represents a real time block that corresponds to a discretized version of a duration time. Finally, we zero pad $s(T_{a,m})$ and $dt(T_{a,m})$ to the maximum observed lengths in our data (obtained by searching over all created sequences over all agents and days), incorporate start, $[BOS]$, separating, $[SEP]$, and end, $[EOS]$, tokens, and concatenate them. Thus, the entire data sequence can be encoded as: 
\begin{equation*}
x_{a,m} = [97, 1, 2, 5, 3, 1, ..., 0, 98, 36, 32, 24, 24, 32, ..., 0, 99],
\end{equation*}
where $97$, $98$, and $99$ represent $[BOS]$, $[SEP]$, and $[EOS]$ tokens, respectively. 

\subsection{Visualization of Subpopulations in our Simulated Dataset} \label{app:vis_subpop}

\begin{figure}[H]
 \centering
  \subfloat{\includegraphics[width=0.8\linewidth]{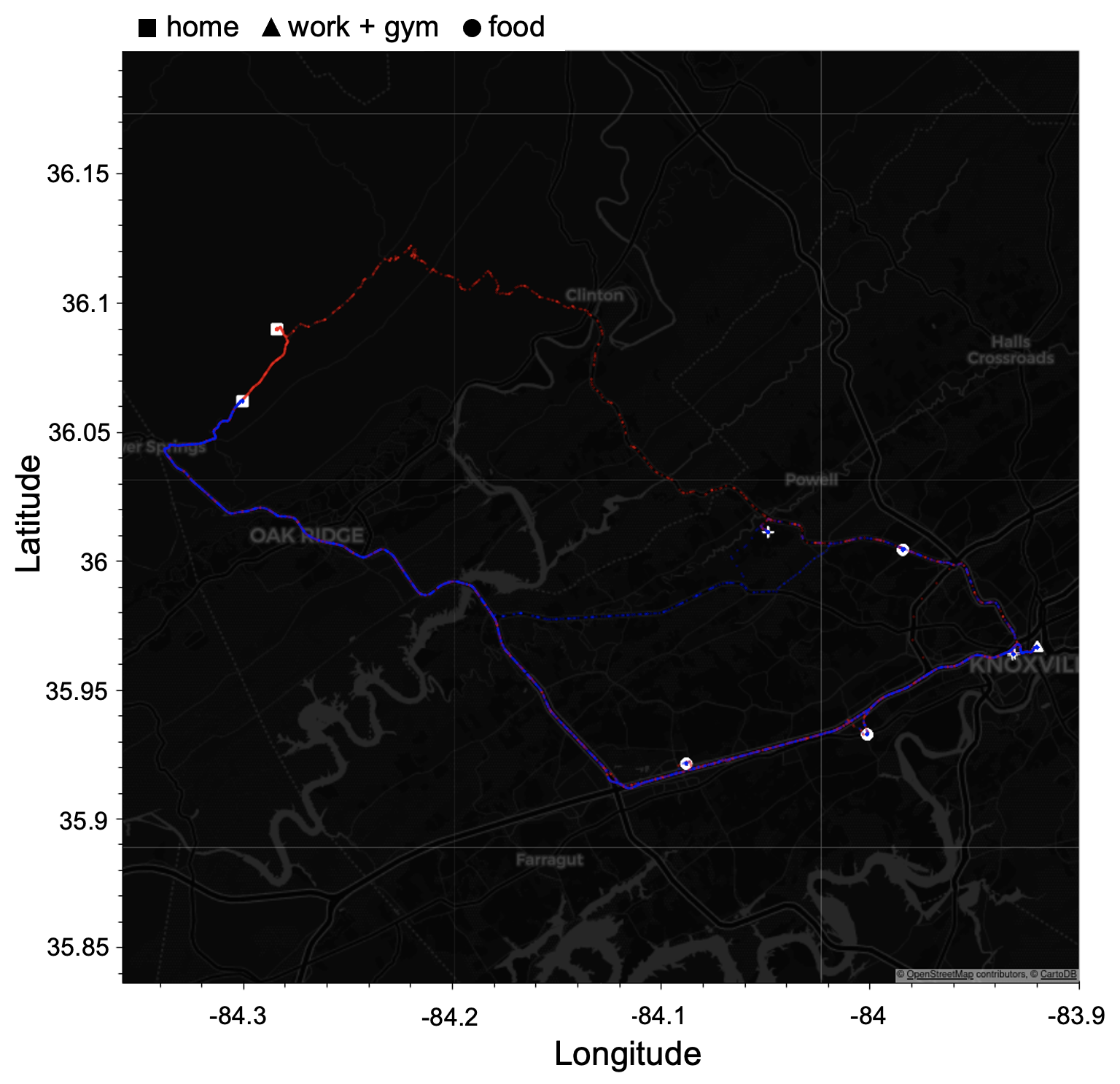}}
  \newline
  \subfloat{\includegraphics[width=0.85\linewidth]{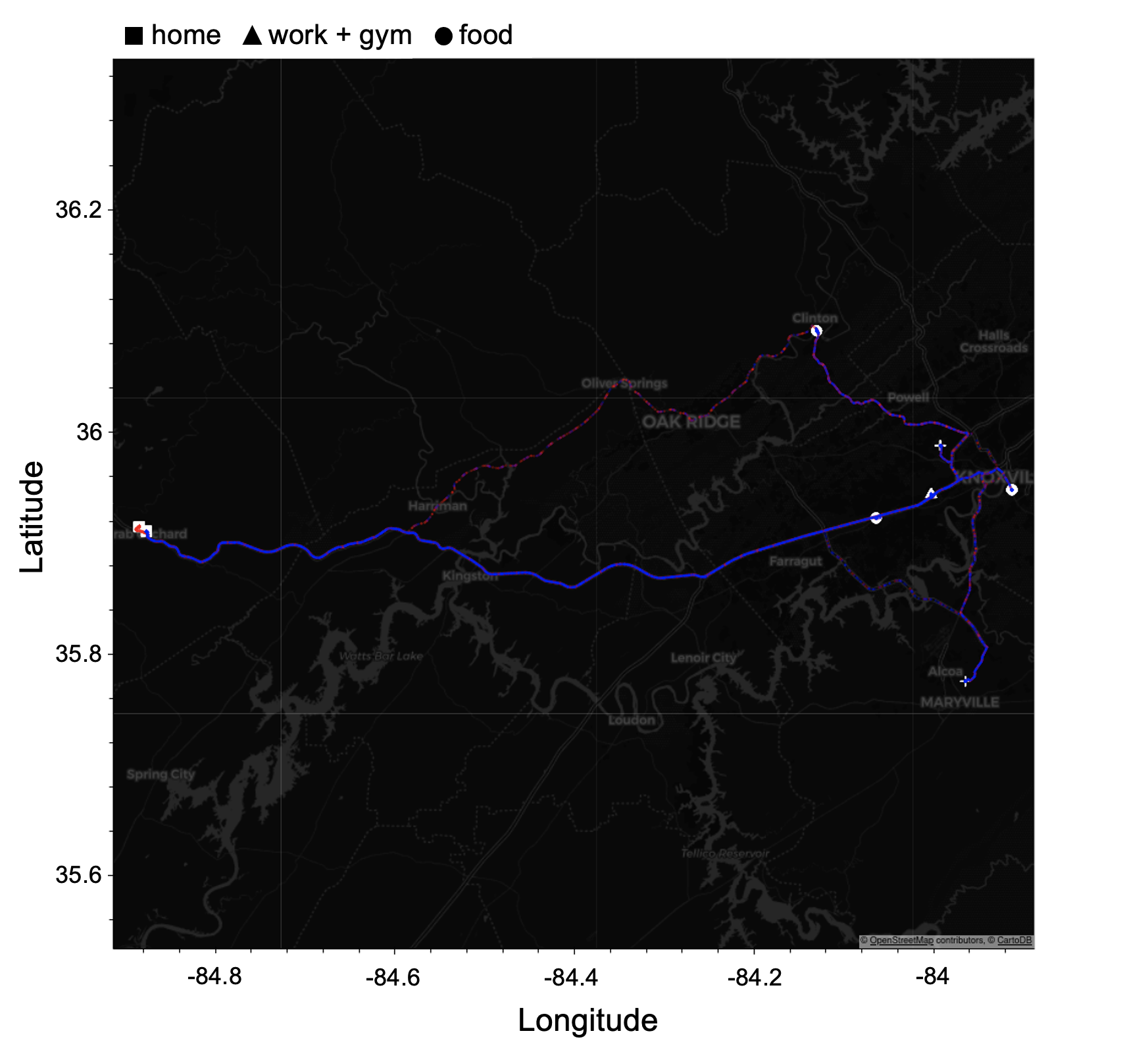}}
  \caption{Visualization of two agents in two different subpopulations of our simulated dataset.}
  \label{fig:4subpops}
\end{figure}

Figure \ref{fig:4subpops} visualizes two agents in two different subpopulations of our simulated data. In the top plot, the two agents (shown in different colors) share many of the same locations, but their trajectories are slightly different due to their differing home locations. In the bottom plot, the two agents have much closer home locations and live in a far more rural area away from the city center so they often take the same major highway into the city to all the commercial locations (work places, restaurants, gyms).

\subsection{Additional Experimental Analysis} \label{app:exp}

\subsubsection{Simulated Data Results}\label{app:sim_1}

Here, we provide interpretations of the results from the blue and red rectangles displayed in Figure \ref{fig:masked_conf_labels}. 

\begin{figure}[h]
    \centering
    \includegraphics[width=0.84\linewidth]{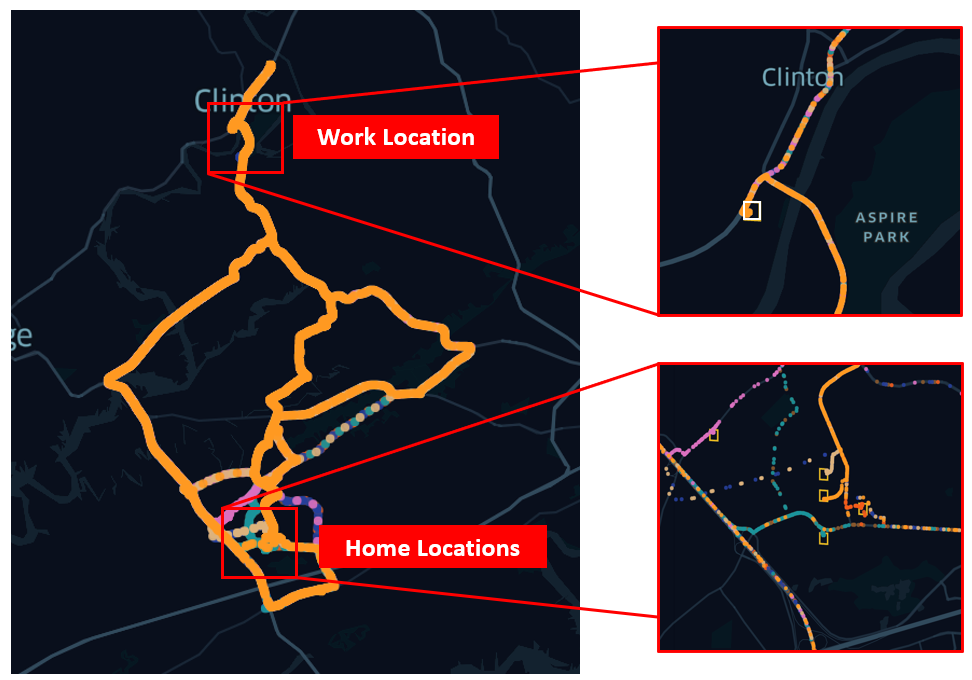}
    \caption{Visualization of the PoLs of agents residing in the red square in Figure \ref{fig:masked_conf_labels} which shows similarly predicted S2 cells containing homes of agents in the same subpopulation.}
\label{fig:masked_red_visual}
\end{figure}

Figure \ref{fig:masked_red_visual} focuses on the red rectangle where home locations get misclassified.  We first plot the PoLs of agents that have PLs that overlap with the misclassified ones in the rectangle of interest, and we see a similar result to that shown in Figure \ref{fig:masked_green_visual} where these agents share the same work location but have different home locations.  Thus, these home locations understandably get misclassified.  

\begin{figure}[h]
    \centering
    \includegraphics[width=0.84\linewidth]{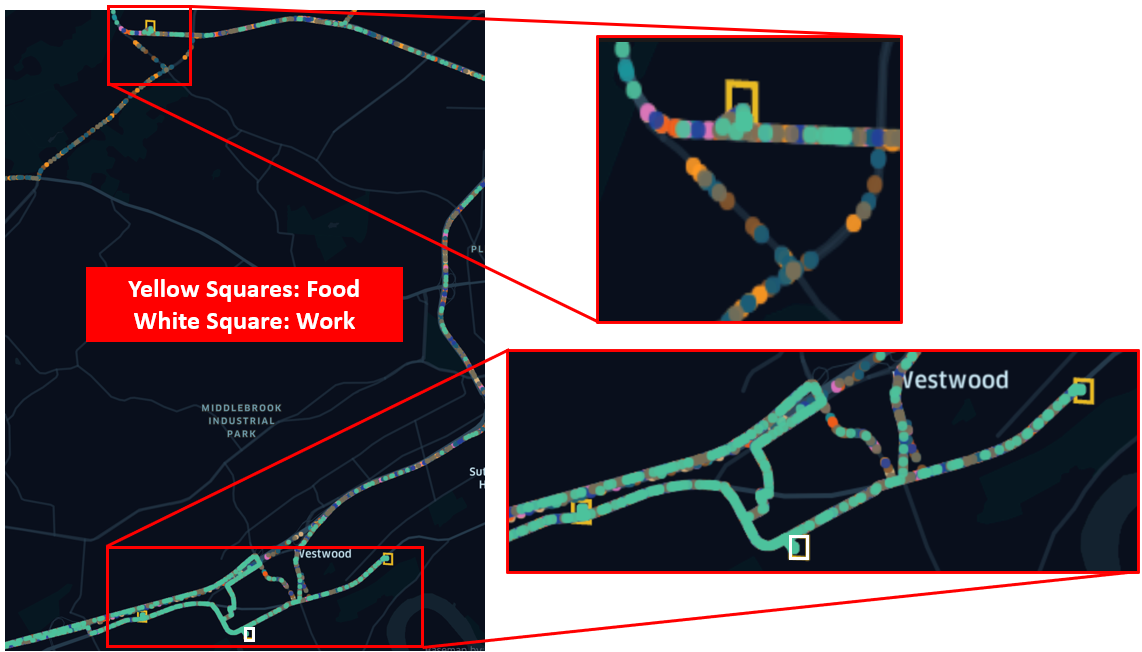}
    \caption{Visualization of the PoLs of agents residing in the blue square in Figure \ref{fig:masked_conf_labels} which shows similarly predicted S2 cells containing the food locations frequented by agents in the same subpopulation.}
\label{fig:masked_blue_visual}
\end{figure}

Figure \ref{fig:masked_blue_visual} focuses on the blue rectangle where food locations get misclassified.  We first plot the PoLs of agents that have PLs that overlap with the misclassified ones in the blue rectangle. We see that these agents share the same work location, but they have different food locations in their PoLs.  In the PoLs, agents can go to different food locations on different days. These agents also share the same work place, further tying the similarity of the food locations. 

\vspace{-6mm}

\subsubsection{``Massive Trajectory Data Based on Patterns of Life" Results}\label{app:sim_2}

\begin{figure}[H]
    \centering
    \includegraphics[width=0.6\linewidth]{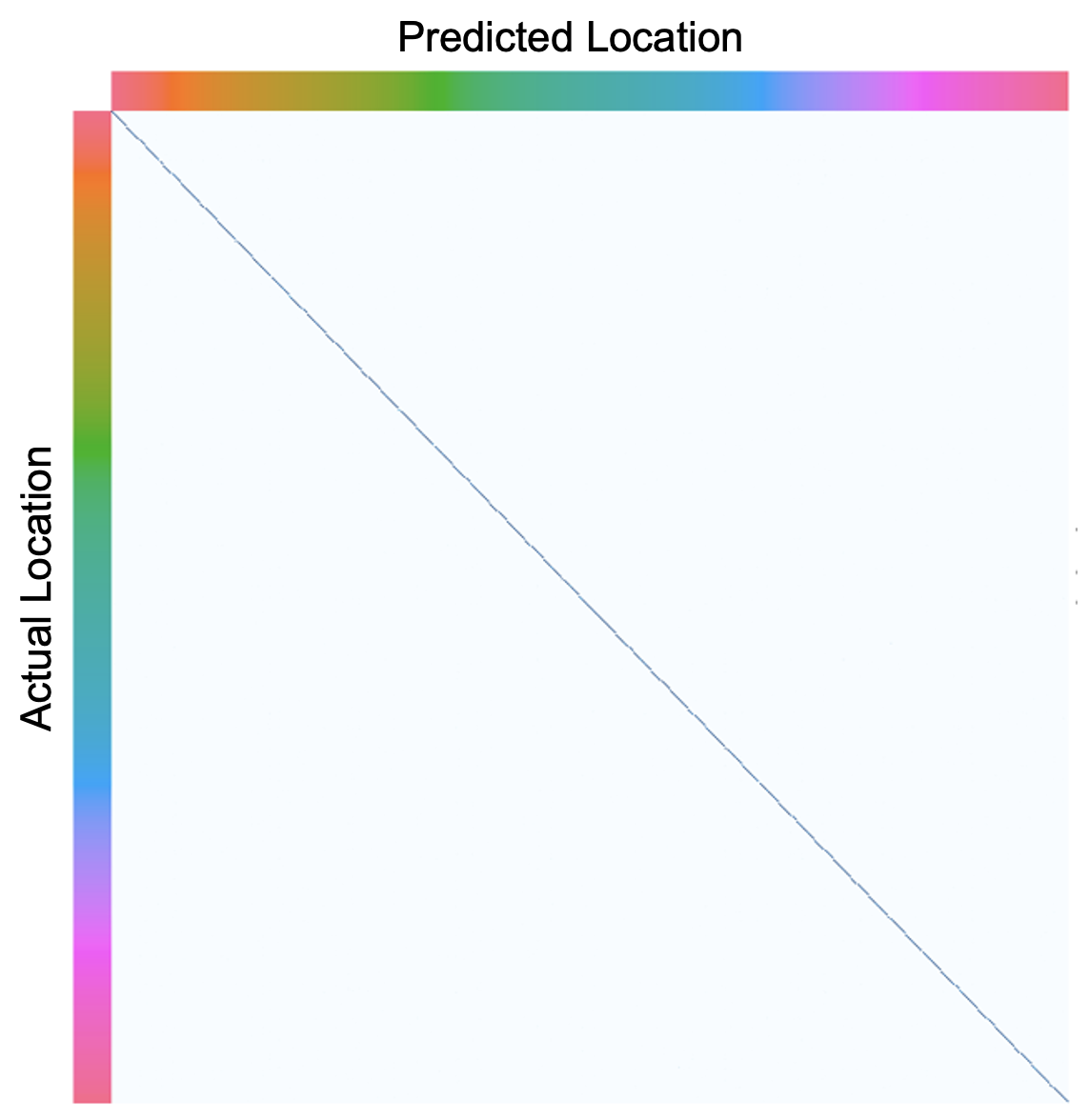}
    \caption{Matrix of average predicted probability scores for the ATL dataset where the rows are the S2 cell locations and the columns are the predicted locations.}
    \label{fig:atlanta_masked_matrix}
\end{figure}

\begin{figure}[H]
    \centering
        \includegraphics[width=0.49\linewidth]{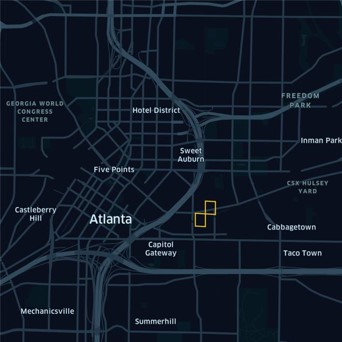}
    \hfill
        \includegraphics[width=0.49\linewidth]{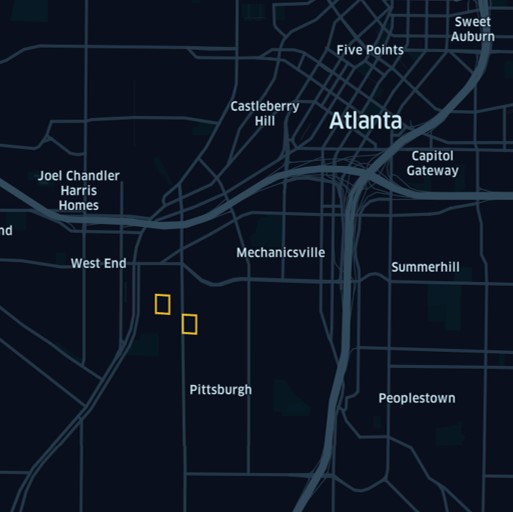}
    \caption{Similar locations learned by STARE when training with the Masked Location Modelling (MLM) task.}
    \label{fig:atl_masked}
\end{figure}

Figure \ref{fig:atlanta_masked_matrix} displays the average predicted post-softmax probability matrix of the Atlanta (ATL) dataset.  We see that locations in general are predicted correctly.  When we plot out some of the higher valued off-diagonals, which represent misclassifications, we get Figure \ref{fig:atl_masked} where we see that the mislabeled locations are actually proximal locations.  Due to this proximity, it is understandable that these locations can be misclassified.

\subsubsection{Ravens Results} \label{app:ravens}

\begin{figure}[H]
    \centering
    \includegraphics[width=0.6\linewidth]{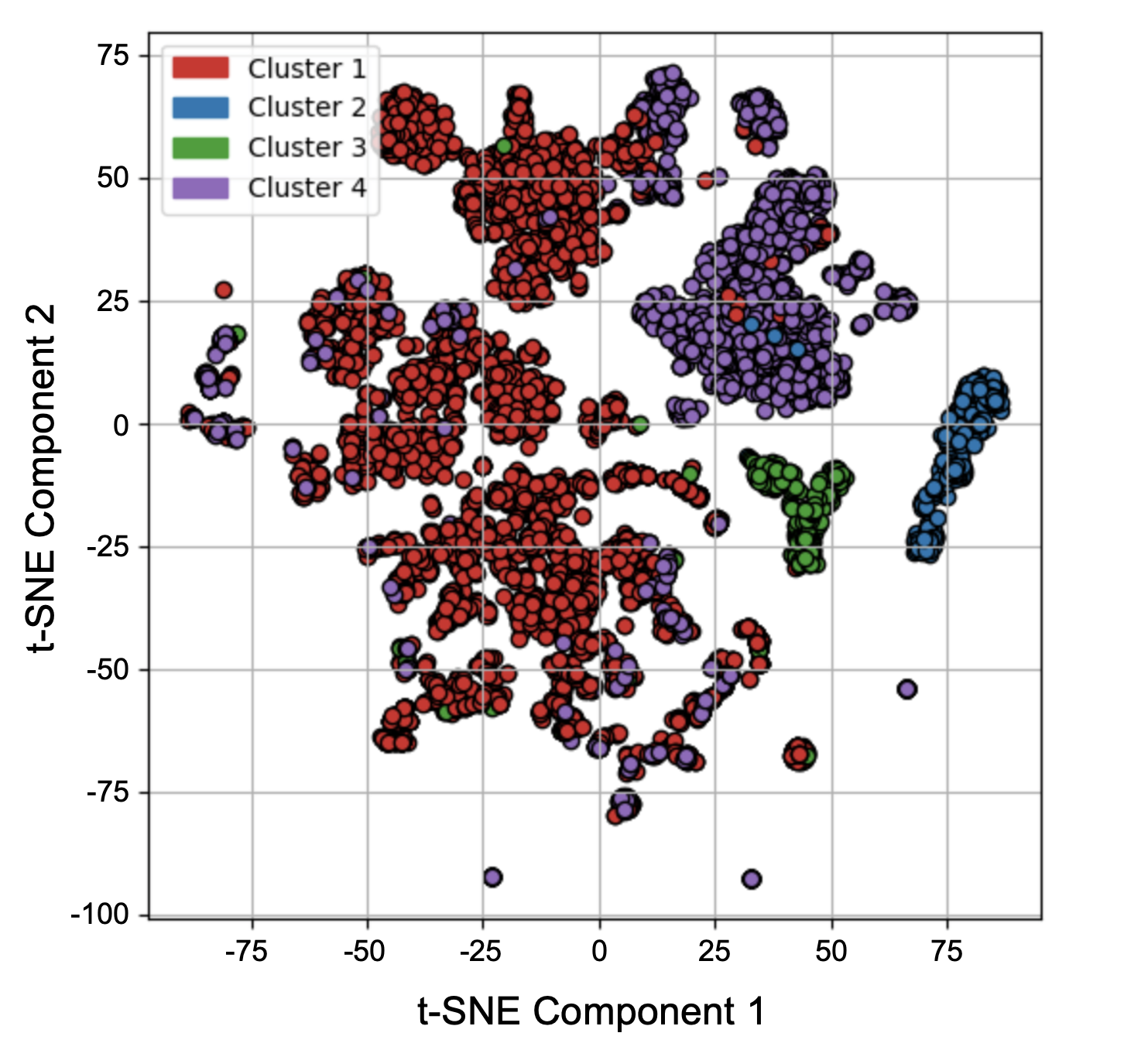}
    \vspace{-1mm}
    \caption{Ravens t-SNE embeddings; the data points are colored based on Figure \ref{fig:raven_spectral_clusters}'s cluster membership.}
    \label{fig:raven_tsne}
\end{figure}

Figure \ref{fig:raven_tsne} displays $2$D t-SNE embeddings of the TES data encodings for the Ravens dataset. All the data points are colored based on the cluster membership defined in Figure \ref{fig:raven_spectral_clusters}; we see that the $4$ clusters obtained via spectral clustering are fairly separated in the embedding space, thus further giving us confidence that our STARE model is able to obtain meaningful embeddings that can be used to learn relationships between agents.  

\end{document}